\documentclass[lettersize,journal]{IEEEtran}
\usepackage{amsmath,amsfonts}
\usepackage{algorithmic}
\usepackage{algorithm}
\usepackage{array}
\usepackage[caption=false,font=normalsize,labelfont=sf,textfont=sf]{subfig}
\usepackage{textcomp}
\usepackage{stfloats}
\usepackage{url}
\usepackage{verbatim}
\usepackage{graphicx}
\usepackage{cite}
\usepackage{booktabs}       
\usepackage{amsfonts}       
\usepackage{nicefrac}       
\usepackage{xcolor}         
\usepackage{graphicx}
\usepackage{amsmath}
\usepackage{algorithm}
\usepackage{algorithmic}
\usepackage{wrapfig}
\usepackage{subfig,graphicx}
\usepackage{multirow}
\usepackage{amssymb}
\usepackage{wrapfig}
\usepackage{hyperref}

\begin{document}

\title{A Simple Unified Uncertainty-Guided Framework for Offline-to-Online Reinforcement Learning}

\author{
Siyuan Guo, Yanchao Sun, Jifeng Hu, Sili Huang, Hechang Chen, \\ Haiyin Piao, Lichao Sun, and Yi Chang,~\IEEEmembership{Senior Member, ~IEEE}

\IEEEcompsocitemizethanks{
\IEEEcompsocthanksitem This work is supported by National Key R\&D Program of China under Grant (2023YFF0905400), National Natural Science Foundation of China through grants (U2341229,
61976102, U19A2065, 62476110), Key R\&D Project of Jilin Province
(20240304200SF), and International Cooperation Project of Jilin
Province (20220402009GH). Hechang Chen and Yi Chang are the corresponding authors.
\IEEEcompsocthanksitem Siyuan Guo, Jifeng Hu, Sili Huang, Hechang Chen and Yi Chang are with the School of Artificial Intelligence, Jilin University, China and Engineering Research Center of Knowledge-Driven Human-Machine Intelligence, MOE, China. Siyuan Guo and Yi Chang are also with the International Center of Future Science, Jilin University, China. \\ E-mail: guosyjlu@gmail.com, \{hujf21, huangsl21\} @mails.jlu.edu.cn, chenhc@jlu.edu.cn, yichang@jlu.edu.cn.
\IEEEcompsocthanksitem Yanchao Sun is with the JPMorgan AI Research, USA. \\ E-mail: ycs@umd.edu
\IEEEcompsocthanksitem Haiyin Piao is with the Northwestern Polytechnical University, China. \\ E-mail: haiyinpiao@mail.nwpu.edu.cn
\IEEEcompsocthanksitem Lichao Sun is with the Lehigh University, USA. \\ E-mail: lis221@lehigh.edu
}
}

\markboth{Journal of \LaTeX\ Class Files,~Vol.~14, No.~8, August~2021}%
{Shell \MakeLowercase{\textit{et al.}}: A Sample Article Using IEEEtran.cls for IEEE Journals}


\maketitle

\begin{abstract}
Offline reinforcement learning (RL) provides a promising solution to learning an agent fully relying on a data-driven paradigm. However, constrained by the limited quality of the offline dataset, its performance is often sub-optimal. Therefore, it is desired to further finetune the agent via extra online interactions before deployment. Unfortunately, offline-to-online RL can be challenging due to two main challenges: \textit{constrained exploratory behavior} and \textit{state-action distribution shift}. In view of this, we propose a \textbf{S}imple \textbf{U}nified u\textbf{N}certainty-\textbf{G}uided (SUNG) framework, which naturally unifies the solution to both challenges with the tool of uncertainty. Specifically, SUNG quantifies uncertainty via a VAE-based state-action visitation density estimator. To facilitate efficient exploration, SUNG presents a practical optimistic exploration strategy to select informative actions with both high value and high uncertainty. Moreover, SUNG develops an adaptive exploitation method by applying conservative offline RL objectives to high-uncertainty samples and standard online RL objectives to low-uncertainty samples to smoothly bridge offline and online stages. SUNG achieves state-of-the-art online finetuning performance when combined with different offline RL methods, across various environments and datasets in D4RL benchmark. Codes are made publicly available in \url{https://github.com/guosyjlu/SUNG}.
\end{abstract}


\section{Introduction}
Offline reinforcement learning (RL) \cite{offline-rl-survey, offline-RL-tnnls}, which enables agents to learn from a fixed dataset without interacting with the environment, has demonstrated significant success in many tasks where interaction is expensive or risky \cite{robot, clinical, dt-recommender}.  However, such a data-driven learning paradigm inherently limits the agent's performance as it fully relies on a typically sub-optimal dataset \cite{br, pex}. To overcome this problem, the RL community has been exploring the offline-to-online setting, which incorporates additional online interactions to further finetune the pretrained offline RL agents \cite{br,pex,odt}.

While the pretraining-finetuning paradigm in offline-to-online RL is natural and intuitive, delivering the expected online improvement can be challenging in practice due to two main challenges: (1) \textit{Constrained exploratory behavior}: The conservative offline RL objectives, which require the agents to perform actions within the support of the dataset during pretraining, can constrain the agents' exploratory behavior for online interactions. As a result, they cannot achieve efficient exploration, and thus fail to fully benefit from the trial-and-error paradigm in online RL. (2) \textit{State-action distribution shift}: During finetuning stage, agents may encounter unfamiliar state-action regimes that fall outside of the support of the dataset. Accordingly, the state-action distribution shift between offline and online data occurs, leading to the well-known extrapolation error \cite{bcq, cql} during exploitation, which may wipe out the good initialization obtained from the pretraining stage. Existing research typically focuses on tackling one aspect of the challenges above by developing either efficient exploration \cite{odt, o3f} or effective exploitation \cite{br,apl,pex}, resulting in limited offline-to-online improvement. However, simultaneously and efficiently addressing both challenges brings additional difficulty: \textit{aggressive exploration may exacerbate the distribution shift, while conservative exploitation can hinder agents from efficient online finetuning.}

In this work, we aim to provide a unified solution to both challenges for better sample efficiency during online finetuning. To achieve this, we recognize that both challenges are closely tied to appropriate identification and treatment of unseen state-action pairs. Specifically, efficient exploration necessitates discovery of informative unseen regions of state-action space. Furthermore, effective exploitation needs precise characterization of out-of-distribution (OOD) data to avoid either overly conservative or aggressive policy learning. As such, we leverage proper quantification and utilization of the uncertainty \cite{sunrise,pbrl,edac,uwac} to naturally address both challenges above and achieve the desired trade-off between exploration and exploitation.

To this end, we present a \textbf{S}imple \textbf{U}nified \footnote{Although our framework can be applied to different offline RL algorithms, we still need to manually adjust a few configurations to accommodate different backbone algorithms, which we detail in Section~\ref{sec:exploration}.} u\textbf{N}certainty \textbf{G}uided (SUNG) framework, which provides a generic solution for offline-to-online RL to enable finetuning agents pretrained with different offline RL objectives. Unlike recent uncertainty-aware RL methods that rely on the ensemble technique for uncertainty quantification \cite{pbrl,edac, sunrise}, SUNG utilizes a simple yet effective approach that estimates the state-action visitation density with a variational auto-encoder (VAE) \cite{vae} to quantify the state-action uncertainty. In contrast to prior offline-to-online RL methods, SUNG simultaneously addresses both challenges by leveraging the tool of uncertainty. Concretely, we develop a practical optimistic exploration strategy that follows the principle of optimism in the face of uncertainty \cite{r-max, ucb}. The main idea here is to select those state-action pairs with both high value and high uncertainty for efficient exploration. We also propose an adaptive exploitation method to handle the state-action distribution shift by identifying and constraining OOD samples. The key insight is to leverage conservative offline RL objectives for high-uncertainty samples, and standard online RL objectives for low-uncertainty samples. This enables agents to smoothly adapt to changes in the state-action distribution. Moreover, we utilize offline-to-online replay buffer (OORB) \cite{apl} to incorporate offline and online data during finetuning. 

Notably, the insights of SUNG are generally applicable and can be combined with most model-free offline RL methods in principle. Empirically, SUNG benefits from consideration of uncertainty to guide both exploration and exploitation, thereby exceeding state-of-the-art methods on D4RL benchmarks \cite{d4rl}, with 14.54\% and 14.93\% extra averaged offline-to-online improvement compared to the best baseline in MuJoCo and AntMaze domains, respectively.
In addition, SUNG demonstrates remarkable robustness in finetuning performance across a range of hyper-parameters.
Furthermore, we showcase SUNG's seamless integration with advanced techniques, such as ensemble Q-learning \cite{br,e2o} and high update-to-data (UTD) ratio \cite{onlineRLwithOfflineData}, leading to superior finetuning performance.

In summary, our main contributions are four-fold:
\begin{itemize}
    \item We propose a generic framework SUNG for sample-efficient offline-to-online RL, flexibly integrating with existing offline RL methods.
    \item We introduce an optimistic exploration strategy via bi-level action selection to select informative actions for efficient exploration.
    \item We develop an adaptive exploitation method with OOD sample identification and regularization to smoothly bridge offline RL and online RL objectives.
    \item Experimental results demonstrate that SUNG outperforms the previous state-of-the-art when combined with different offline RL methods, across various types of environments and datasets.
\end{itemize}

\section{Related Work}
\label{sec:related_work}
\subsection{Offline RL} 
Offline RL \cite{offline-rl-survey} aims at learning a policy solely from a fixed offline dataset. Most prior works in offline RL have been dedicated to addressing the extrapolation error due to querying the value function with OOD actions \cite{bcq, cql, policy-regularization, policy-regularization-2, value-based-regularization}. As such, it is critical to constrain the learned policy to perform actions within the support set of the offline dataset. Common strategies include policy constraint \cite{bcq,bear,td3+bc,spot}, value regularization \cite{cql, mcq, fisherbrc, pbrl}, etc. However, previous works have shown that the the performance of offline RL agent is limited by the quality of the datasets \cite{awac, br}, which motivates investigation for the offline-to-online setting.

\subsection{Offline-to-Online RL} 
Offline-to-online RL involves pretraining with offline RL and finetuning via online interactions. Some offline RL methods naturally support further online finetuning with continual offline RL objectives \cite{spot,mcq,iql}. However, they typically tend to be too conservative and yield limited performance improvement \cite{aca}. Besides, some offline-to-online RL approaches are designed for one specific offline RL method \cite{cal-QL,td3+bc-o2o,oema}. For example, ODT \cite{odt} introduces the max-entropy RL for finetuning DT \cite{dt}. ABCR \cite{abcr} proposes to adaptively loosen the conservative objectives of TD3+BC \cite{td3+bc}. 
In contrast, we focus on the generic offline-to-online RL framework that can be combined with different offline RL methods \cite{br,o3f,apl,pex,proto,e2o}. Prior works typically address the challenges of exploration limitation and state-action distribution shift. For the former, O3F \cite{o3f} utilizes the knowledge in value functions to guide exploration, but it is not compatible with value regularization based offline RL methods. For the latter, BR \cite{br} utilizes the ensemble technique together with a balanced replay buffer that prioritizes near-on-policy samples. APL \cite{apl} proposes to take different advantages of offline data and online data for adaptive policy learning. PEX \cite{pex} freezes the pretrained policies and trains a new policy from scratch using an adaptive exploration strategy to avoid erasing pre-trained policies. Unlike them, SUNG unifies both challenges by emphasizing the proper estimation and utilization of uncertainty.

\subsection{Uncertainty for RL} Uncertainty-aware RL has achieved notable success in reinforcement learning \cite{url-survey}, with uncertainty typically categorized as either aleatoric or epistemic. Aleatoric uncertainty arises from the inherent stochasticity of the environment and agent interactions. Distributional RL \cite{c51, qr-dqn, iqn} models this by predicting the return distribution, rather than the exact Q-value. In contrast, epistemic uncertainty originates from the insufficient knowledge and data of the environment, where Bayesian uncertainty quantification methods are widely applied. For example, ensembling based methods \cite{sunrise, edac} utilize the variance of predictions between different models to quantify the uncertainty. Bootstrapped learning based methods \cite{pbrl,bdqn} extend this idea by training these models on different resampled subsets of the data, further enhancing diversity and robustness. Similarly, Monte-Carlo dropout based methods \cite{bdropout,uwac} estimate uncertainty by performing multiple stochastic forward passes through a network with dropout enabled at inference time, and computing the variance across these outputs. Recently, generative models, such as VAE \cite{vae, cpq} and diffusion models \cite{diffusion-1, diffusion-2}, are also applied for uncertainty quantification. Different from the aforementioned works, we focus on the offline-to-online RL setting, and we discuss detailed differences from them in Appendix A in the supplementary material.


\section{Preliminaries}
\label{sec:preliminary}
\noindent\textbf{RL.} 
We follow the standard RL setup that formulates the environment as a Markov decision process (MDP) $\mathcal{M}=(\mathcal{S}, \mathcal{A}, p, r, \gamma)$, where $\mathcal{S}$ is the state space, $\mathcal{A}$ is the action space, $p(s'|s,a)$ is the transition distribution, $r(s,a)$ is the reward function, and $\gamma \in [0,1)$ is the discount factor. The goal of RL is to find a policy $\pi(a|s)$ that maximizes the expected return $\mathbb{E}_\pi[\sum_{t=0}^{\infty}\gamma^tr(s_t, a_t)]$. We measure this objective by a state-action value function 
\begin{equation}
    Q^\pi(s,a)=\mathbb{E}_\pi\left[\sum_{t=0}^{\infty}{\gamma^tr_t}|s_0=s,a_0=a\right],
\end{equation}
which calculates the expected discounted return after taking the action $a$ in state $s$.

\noindent\textbf{Off-policy RL.}
Off-policy RL methods, such as TD3 \cite{td3} and SAC \cite{sac}, have been widely applied due to their sample efficiency. These methods typically alternate between policy evaluation and policy improvement. In particular, given an experience replay dataset $\mathcal{D}$, TD3 learns a deterministic policy $\pi_\phi(s)$ and a state-action value function $Q_\theta(s,a)$, parameterized by $\phi$ and $\theta$, respectively. The value function can be updated via temporal difference (TD) learning as
\begin{equation}
\label{eqn:q}
    \mathcal{L}_Q(\theta) = \mathbb{E}_{(s,a,r,s')\sim \mathcal{D}} \left[\left(Q_\theta(s,a) -r -\gamma Q_{\bar{\theta}} \left(s',\pi_\phi\left(s'\right)\right)\right)^2\right],
\end{equation}
where $Q_{\bar{\theta}}$ is the target value network for stabilizing the learning process. Then, the policy can be updated to maximize the current Q value:
\begin{equation}
\label{eqn:pi}
    \mathcal{L}_\pi(\phi) = \mathbb{E}_{s \sim \mathcal{D}}\left[ -Q_\theta\left(s,\pi_\phi\left(s\right)\right)\right].
\end{equation}

\noindent\textbf{Offline RL.}
In the offline RL setting, the agent only has access to a fixed dataset $\mathcal{D}=\{(s,a,r,s')\}$. Although off-policy RL methods can learn from data collected by any policy in principle, they fail in the offline RL setting. This can be attributed to the well-known extrapolation error \cite{bcq, cql} due to querying the value function with OOD actions. As such, one main line of model-free offline RL research is to constrain the learned policy to perform actions within the support of the dataset in different ways, such as policy constraint \cite{td3+bc, spot}, value regularization \cite{cql, fisherbrc, mcq} and etc. Among them, two representative offline RL methods are TD3+BC \cite{td3+bc} and CQL \cite{cql}. The former adds a behavior cloning (BC) regularization term to the standard policy improvement in TD3:
\begin{equation}
\label{eqn:td3+bc}
    \mathcal{L}_{\pi}^{\rm{TD3+BC}}(\phi) = \mathcal{L}_\pi(\phi) + \lambda_{\rm{BC}}\mathbb{E}_{(s,a) \sim \mathcal{D}}\left[ \left(\pi_\phi\left(s\right)-a\right)^2\right],
\end{equation}
where $\lambda_{\rm{BC}}$ balances the standard policy improvement loss and BC regularization. In contrast, the latter resorts to pessimistic under-estimate of Q values during policy evaluation in SAC:

\begin{equation}
\begin{aligned}
    &\mathcal{L}_{Q}^{\rm{CQL}}(\theta) = \mathcal{L}_Q(\theta) \\ &+\lambda_{\rm{CQL}}\left(\mathbb{E}_{s\sim \mathcal{D},a\sim\pi_\phi}\left[Q_\theta\left(s,a\right)\right] + \mathbb{E}_{(s,a)\sim\mathcal{D}}\left[-Q_\theta\left(s,a\right)\right]\right),
\end{aligned}
\label{eqn:cql}
\end{equation}
where $\lambda_{\rm{CQL}}$ denotes the trade-off factor. Then, we can formally summarize both methods as 
\begin{equation}
\label{formalization}
    \mathcal{L}_{\mathcal{F}}^{\rm{offline}}(\Theta) = \mathcal{L}_{\mathcal{F}}(\Theta) + \lambda \mathcal{R}_{\mathcal{F}}(\Theta),
\end{equation}
where $\mathcal{F}$ represents either a policy or a value function, parameterized by $\Theta$, $\mathcal{R}$ denotes a regularizer to prevent the extrapolation error, and $\lambda$ balances standard loss and regularization. Note that most model-free offline RL methods can be summarized as Eq. (\ref{formalization}) in an explicit \cite{cql, td3+bc, fisherbrc, spot, mcq} or implicit manner \cite{bear, anti-exploration, iql, pbrl, ivr}, which motivates the adaptive exploitation in Section \ref{section:adaptive-exploitation}.

\noindent\textbf{Offline-to-Online RL.}
While offline RL has found widespread application in constrained interaction scenarios, its efficacy is inherently tied to the quality of the offline dataset. Thus, it becomes imperative to engage in subsequent online finetuning before deploying the model. Formally, the concept of offline-to-online RL delves into the study of a scenario where the agent $\pi_\phi$, often accompanied by the Q function $Q_\theta$, undergoes initial pretraining using the offline dataset $\mathcal{D}$ through a designated offline RL algorithm. Following this pretraining phase, the agent is further finetuned via online interactions with the environment. This online finetuning process employs off-policy RL algorithms to align the training methodology with that employed during offline pretraining. Additionally, the online replay buffer $\mathcal{B}$ is typically initialized with the offline dataset $\mathcal{D}$ to facilitate effective reuse of experiences.

\begin{figure*}
    \begin{center}
    \centerline{\includegraphics[width=\textwidth]{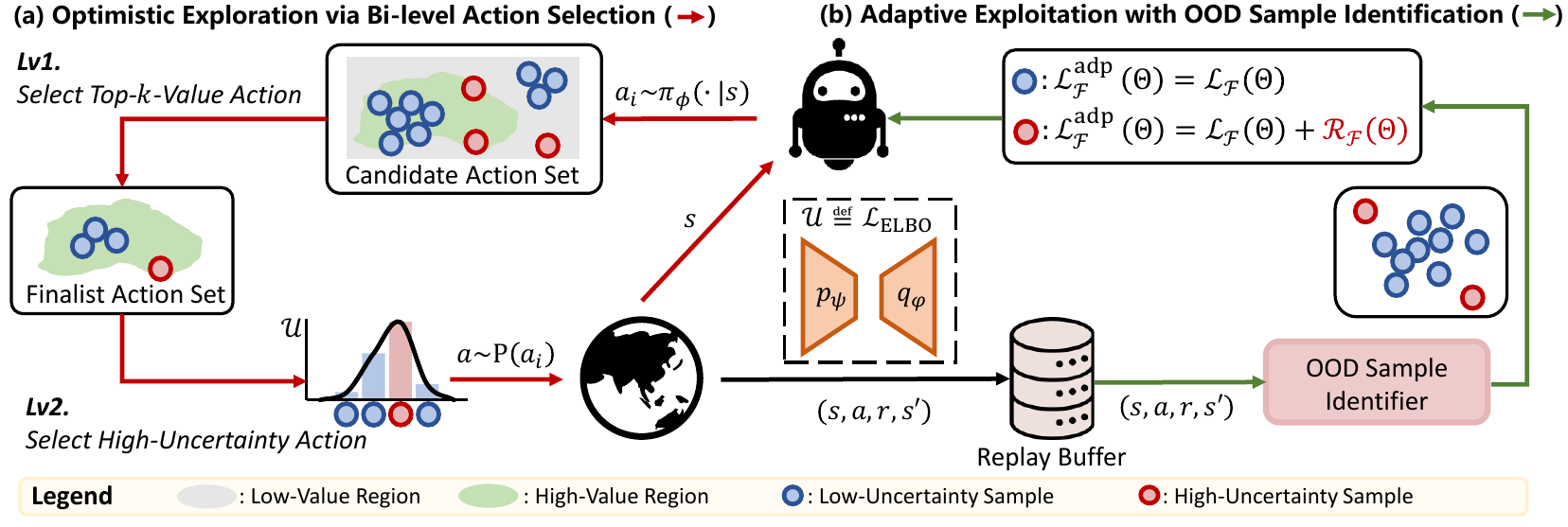}}
    \caption{Overview of SUNG. During online finetuning, we alternate between \textbf{(a)} optimistic exploration strategy to collect behavior data from environment and \textbf{(b)} adaptive exploitation method to improve the policy. We adopt VAE for state-action density estimation to quantify uncertainty. }
    \label{fig:overview}
    \end{center}
\end{figure*}

\section{SUNG}
\label{sec:method}
In this section, we present a simple unified uncertainty-guided framework (SUNG) for offline-to-online RL, as depicted in Fig. \ref{fig:overview}. Concretely, SUNG quantifies the state-action uncertainty by estimating a density function with VAE. Then, under the guidance of uncertainty, SUNG incorporates an optimistic exploration strategy and an adaptive exploitation method for online finetuning. Finally, we summarize the full framework in Algorithm \ref{alg:SUNG}.

\subsection{Uncertainty Quantification with Density Estimation}
Many prior works typically utilize Q ensembles to qualify uncertainty \cite{uwac, o3f, oac, sunrise}. However, the ensemble technique may significantly increase computational costs. Thus, in this work, we utilize a simple yet effective approach by adopting VAE \cite{vae} as a state-action visitation density estimator for uncertainty quantification. Specifically, VAE consists of an encoder $q_\varphi(z|s,a)$ and a decoder $p_\psi(s,a|z)$ with parameters $\varphi$ and $\psi$, respectively.
The VAE is optimized with the evidence lower bound (ELBO):
\begin{equation}
\label{eqn:elbo}
\begin{aligned}
\mathcal{L}_{\mathrm{ELBO}}(s,a;\psi,\varphi) &= \mathbb{E}_{q_\varphi(z|s,a)}\left[-\log p_\psi(s,a|z)\right] \\ &+ D_{\mathrm{KL}}\left[q_\varphi(z|s,a)||p(z)\right],
\end{aligned}
\end{equation}
where $p(z)=\mathcal{N}(0,I)$ is a fixed prior distribution. In ELBO, the first term is the reconstruction loss, while the second term is the KL divergence that regularizes the posterior distribution $q_\varphi(z|s,a)$ to be close to the prior $p(z)$.

After training a VAE based on the offline dataset $\mathcal{D}$, we utilize the negative log likelihood of the state-action density for uncertainty quantification, which can be further approximated by ELBO, i.e., 
\begin{equation}
\begin{aligned}
   \mathcal{U}(s,a)&\overset{\rm{def}}{=}-\log p(s,a) \\
   &=\mathcal{L}_{\mathrm{ELBO}}(s,a;\psi,\varphi)-D_{\mathrm{KL}}[q_\varphi(z|s,a)||p_\psi(z|s,a)]\\
   &\approx\mathcal{L}_{\mathrm{ELBO}}(s,a;\psi,\varphi).
\end{aligned}
\end{equation}
Intuitively, both the reconstruction loss and KL divergence in ELBO indicate the uncertainty for state-action space, i.e., whether a state-action pair can be well captured by the learned distribution in VAE. We provide more justification of such approximation in Appendix B in the supplementary material. 

Note that VAEs have been widely adopted in previous offline RL works—such as BCQ \cite{bcq}, SPOT \cite{spot}, and CPQ \cite{cpq}—to model the behavior policy $\pi_\beta$ of the offline dataset $\mathcal{D}$. This modeled policy is then used to constrain the learned policy from offline RL, preventing it from deviating from the offline policy distribution too much. Different from them, we adopt VAE to estimate the visitation density of the state-action pair as uncertainty quantification, which can guide optimistic exploration and adaptive exploitation in offline-to-online RL. We present a illustrative comparison of them in the supplementary material.

\subsection{Optimistic Exploration via Bi-level Action Selection}
\label{sec:exploration}
A main line of model-free offline RL methods perform conservative objectives to penalize OOD actions during offline pretraining. However, such conservative objectives inherently limit agents' exploration power, which impedes agents to fully benefit from online finetuning \cite{o3f}. To overcome this limitation, our goal is to properly measure the informativeness of unseen actions, and then select the most informative action for exploration. To achieve this, we propose an optimistic exploration strategy via bi-level action selection mechanism. In this subsection, we first formulate the objective of the optimistic exploration strategy, and then propose a practical bi-level action selection mechanism to find approximate solutions of this optimization objective.

To overcome the constrained exploratory behavior issue, we propose an optimistic exploration strategy based on the principle of optimism in the face of uncertainty, which is theoretically grounded on the upper confidence bound (UCB) algorithm \cite{r-max, ucb} in the bandit setting. The key insight here is that we encourage the agents to select informative actions with both high Q value and high uncertainty.

Previous works typically estimate epistemic uncertainty via Q ensembles and utilize the UCB algorithm to direct the exploration \cite{ucb-rl, oac, sunrise}. Different from them, we extend this idea by utilizing  $\mathcal{U}(s,a)$ for uncertainty quantification. To avoid inaccurate estimates for Q value and uncertainty, we only select near-on-policy actions for exploration. Consequently, the objective of the exploration policy $\pi_E(s)$ is to maximize the informativeness while remaining near-on-policy, i.e.,
\begin{equation}
\label{ucb}
\begin{aligned}
    \pi_{E}(s) = & \arg\max_{a \in \mathcal{A}} Q_\theta(s,a)+\beta\mathcal{U}(s,a), \\ 
    & \mathrm{s.t.} \ ||a-\pi_\phi\left(s\right)||^2 \leq \delta,
\end{aligned}
\end{equation}
where $\beta$ controls the optimism level and $\delta$ controls the on-policyness, given a general assumption of smoothness in the physical transition model \cite{td3}. As such, the derived behavior policy can not only encourage informative actions according to the principal of optimism in the face of uncertainty, but also guarantee the on-policyness to ensure the stability.

However, such an optimization problem is impractical to solve due to two main challenges: (1) It is not straightforward to find optimal action in high-dimensional continuous action space, which is impossible to enumerate exhaustively. (2) It is hard to set a proper value for optimism level $\beta$ because Q value and uncertainty have different value ranges across different tasks, which necessitates task-specific hyper-parameter tuning.

To mitigate both challenges above, we propose a practical bi-level action selection mechanism. The key idea here is to approximate the intractable constrained optimization by breaking it into two simpler stages: first narrowing down a promising subset of near-on-policy candidate actions using one criterion (Q or uncertainty), and then probabilistically selecting among them using the complementary criterion. 

Specifically, we first generate a candidate action set $\mathcal{C}=\{a_i\}_{i=1}^{N}$ with $N$ candidate actions. Here, the action is generated by sampling $a_{i} \sim \pi_\phi(\cdot|s)$ for stochastic policy, while by adding sampled Gaussian noise $a_i=\pi_\phi(s)+\epsilon_i, \epsilon_i\sim\mathcal{N}(0,\delta)$ for deterministic policy. Then, we rank the candidate actions according to the Q-values (or uncertainty) to select top-$k$ candidate actions as a finalist action set $\mathcal{C}_f$. Finally, we construct a categorical distribution according to the uncertainty (or Q values) to select the action from $\mathcal{C}_f$ for interacting with the environment:
\begin{equation}
\label{eqn:prob}
    \mathbb{P}(a_i|s):=\frac{\mathrm{exp}\left(\mathcal{U}(s,a_i)/\alpha\right)}{\sum_j\mathrm{exp}\left(\mathcal{U}(s,a_j)/\alpha\right)}, \forall i \in [1, ..., k],
\end{equation}
where $\alpha$ is the softmax temperature. Intuitively, when we set the ranking criteria for the finalist action set as Q values, if we set $k=1$, the exploration policy would greedily choose high-value actions; if we set $k=N$, the exploration policy would favor those with high uncertainty actions. As such, by altering ranking criteria for the finalist action set and $k$, we can flexibly adjust the preference for high Q value or high uncertainty to achieve the desired trade-off. 

As discussed in \cite{o3f}, value regularization based methods may diverge when preference for Q value is present. To address this, we establish the ranking criteria for the finalist action set as uncertainty for value regularization-based methods and as Q value for other offline RL methods.

The optimistic exploration strategy enables agents to interact the environment with informative actions, thereby boosting the sample efficiency of offline-to-online RL. However, it may also bring negative effects by further increasing state-action distribution shift, as a result of the principle of optimism in the face of uncertainty. To this end, we introduce an adaptive exploitation method in the following subsection, which aims to mitigate the state-action distribution shift.

\subsection{Adaptive Exploitation with OOD Sample Identification}
\label{section:adaptive-exploitation}
Action distribution shift poses a significant challenge for offline RL algorithms \cite{bcq, bear, offline-rl-survey, cql}, as the bootstrapping term in policy evaluation involves actions derived from the learned policy $\pi_\phi$. This can result in the extrapolation error due to querying Q functions with OOD actions, leading to biased policy improvement towards these OOD actions with erroneously high Q values. Note that model-free offline RL methods do not suffer from state distribution shift during training, since policy evaluation only queries Q functions with states present in the offline dataset.

However, during online finetuning, both state and action distribution shifts occur since agents may encounter unfamiliar state-action regimes that fall outside of the support of the dataset. Worse still, as the proposed optimistic exploration strategy follows the principle of optimism in the face of uncertainty, the state-action distribution shift may be further exacerbated. Preliminary experiments from \cite{br, o3f, apl} show that this may erase the good initialization obtained from offline pretraining. 

To tackle the state-action distribution shift issue, we propose an adaptive exploitation method with OOD sample identification. The underlying idea is developed from the observations in previous works \cite{br, o3f, apl} that finetuning with continual offline RL objectives typically derives stable but limited performance, while finetuning with online RL objectives typically derives unstable performance due to the distribution shift caused by OOD state-action pairs. Therefore, we propose a novel approach that leverages the benefits of both objectives by utilizing conservative offline RL objectives for state-action pairs with high uncertainty while aggressive online RL objectives for state-action pairs with low uncertainty. As shown in Eq. (\ref{formalization}), offline RL objectives consist of a standard online RL objective $\mathcal{L}_{\mathcal{F}}(\Theta)$ and a regularizer $\mathcal{R}_{\mathcal{F}}(\Theta)$. Thus, we can derive the objectives for adaptive exploitation by introducing an uncertainty-guided OOD sample identifier $\mathcal{I}(s,a)$: 
\begin{equation}
\label{eqn:adp}
    \mathcal{L}_{\mathcal{F}}^{\rm{adp}}(\Theta) = \mathcal{L}_{\mathcal{F}}(\Theta) + \lambda  \mathcal{I}\left(s, \pi_\phi\left(s\right)\right)\mathcal{R}_{\mathcal{F}}(\Theta).
\end{equation}
We expect that $\mathcal{I}(s,a)$ is a large value for state-action pairs with high uncertainty and a small value for those with low uncertainty. Therefore, we construct such an identifier using uncertainty estimation $\mathcal{U}(s,a)$ as below. For a minibatch of state-action pairs $\{s_i, a_i\}_{i=1}^{M}$, we select the top $p\%$ of them as OOD state-action pair sets $\mathcal{D}_{\mathrm{OOD}}$, according to its corresponding uncertainty estimation $\mathcal{U}(s_i, a_i)$. The selection process can be performed by sampling from a categorical distribution similar to Eq. (\ref{eqn:prob}). Then, out of simplicity, we define the OOD sample identifier as 
\begin{equation}
\label{eqn:I}
    \mathcal{I}(s,a) := 
    \left\{ 
    \begin{array}{ll}
        1, & \mathrm{if} (s,a) \in \mathcal{D}_{\mathrm{OOD}}, \\
        0, & \mathrm{else}.\\
    \end{array}
    \right.
\end{equation}
By setting $p=0$, the training objective is precisely the online RL objective, while setting $p=100$ results in the offline RL objective. Thus, by tuning the value of $p$, the proposed adaptive exploitation method can attain the desired trade-off between performance and stability.

A related work, CPQ \cite{cpq}, also employs a VAE to detect OOD actions and modifies the Bellman updates in Q-learning to mitigate overestimation for such actions. In contrast, SUNG leverages OOD sample identification to adaptively introduce regularization terms for OOD samples, aiming to balance performance and stability during online finetuning.

\begin{figure}[t]
\begin{algorithm}[H]
  \caption{SUNG: A Simple Unified Uncertainty-Guided Framework for Offline-to-Online RL}
  \label{alg:SUNG}
\begin{algorithmic}[1]
  \STATE {\bfseries Input:} Dataset $\mathcal{D}$, Offline RL algorithm $\{ \mathcal{L}_{Q}^{\mathrm{offline}}(\theta), \mathcal{L}_{\pi}^{\mathrm{offline}}(\phi)\}$
  \STATE Initialize Q network $Q_\theta$, policy network $\pi_\phi$, VAE network $p_\psi$ and $q_\varphi$
  \WHILE{in \textit{offline pretraining phase}}
    \STATE Sample minibatch of transitions $\left(s, a, r, s'\right) \sim \mathcal{D}$
    \STATE Update $\theta$ and $\phi$ minimizing $\mathcal{L}_{Q}^{\mathrm{offline}}(\theta)$ and $\mathcal{L}_{\pi}^{\mathrm{offline}}(\phi)$
    \STATE Update $\psi, \varphi$ minimizing $\mathcal{L}_{\mathrm{ELBO}}(s, a; \psi, \varphi)$ in Eq. (\ref{eqn:elbo})
  \ENDWHILE
  \STATE Initialize OORB with $\mathcal{B}_{\mathrm{OORB}}\leftarrow \{\mathcal{B}=\emptyset, \mathcal{D}\}$ 
  \WHILE{in \textit{online finetuning phase}}
    \STATE // {\bfseries Optimistic Exploration}
    \STATE Generate candidate action set $\mathcal{C}=\{a_i\}_{i=1}^{N}$
    \STATE Select actions with top-$k$ Q-value (or uncertainty) as finalist action set $\mathcal{C}_f$
    \STATE Sample the action $a_t \sim \mathbb{P}(a_i|s_t), a_i \in \mathcal{C}_f$ in Eq. (\ref{eqn:prob}) according to the uncertainty (or Q value)
    \STATE Execute the action $a_t$, collect next state $s_{t+1}$ and reward ${r_t}$ from the environment
    \STATE Store transitions $(s_t,a_t,r_t,s_{t+1})$ into $\mathcal{B}$ and $\mathcal{D}$
    
    \STATE // {\bfseries Adaptive Exploitation}
    \STATE Sample minibatch of transitions $\left(s, a, r, s^{\prime}\right) \sim \mathcal{B}_{\mathrm{OORB}}$
    \STATE Update $\theta$ and $\phi$ minimizing $\mathcal{L}_{\mathcal{F}}^{\mathrm{adp}}(\Theta)$ in Eq. (\ref{eqn:adp})
    \STATE Update $\psi, \varphi$ minimizing $\mathcal{L}_{\mathrm{ELBO}}(s, a; \psi, \varphi)$ in Eq. (\ref{eqn:elbo})
  \ENDWHILE
\end{algorithmic}
\end{algorithm}
\end{figure}

\subsection{Algorithm Implementation Details}

\subsubsection{Offline-to-Online Replay Buffer}
Previous studies \cite{br, apl} have highlighted the significance of incorporating both offline and online data during finetuning. Here, we adopt the offline-to-online replay buffer (OORB) proposed in \cite{apl} due to its simplicity. Specifically, OORB consists of two replay buffers: a fully online buffer that collects transitions from online interactions, and an offline buffer that stores transitions from both offline dataset and online interactions. During training, each transition in a minibatch is sampled using a Bernoulli-based source selection strategy: for each transition, we flip a Bernoulli coin with success probability $p_{\text{OORB}}$. If the coin lands heads (i.e., success), we sample a transition uniformly from the online buffer; otherwise, we sample uniformly from the offline buffer. This sampling procedure is repeated independently for every transition in the minibatch.

\subsubsection{SUNG Framework for Offline-to-Online RL}
Summarizing the components outlined above, we present the complete framework in Algorithm \ref{alg:SUNG}. SUNG first pretrains agents and VAE estimator using the offline dataset. Then, it turns to online finetuning with the proposed optimistic exploration strategy and adaptive exploitation method. 


\begin{figure}[!t]
    \centering
    \includegraphics[width=\linewidth]{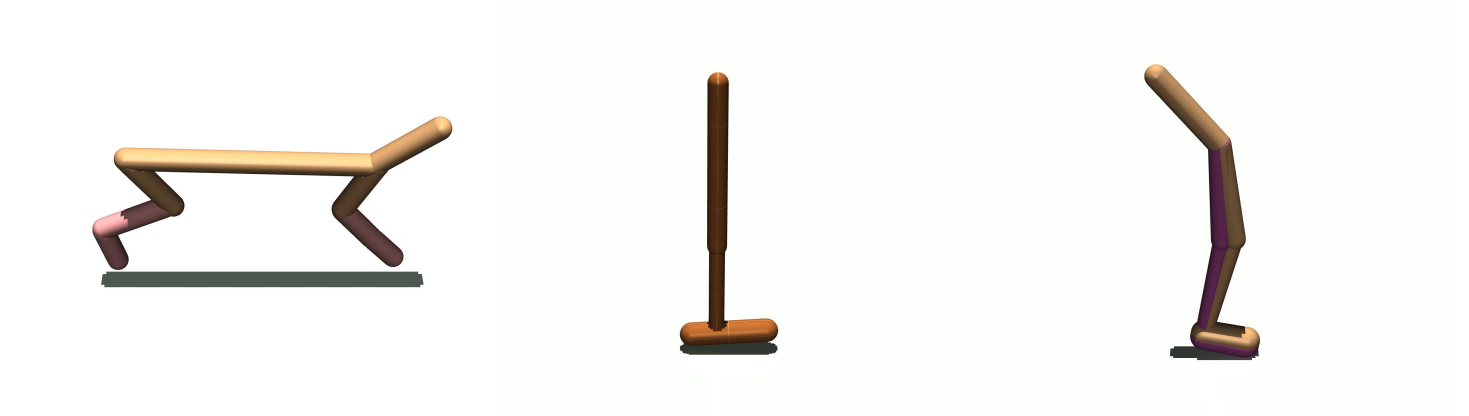}
    \includegraphics[width=0.9\linewidth]{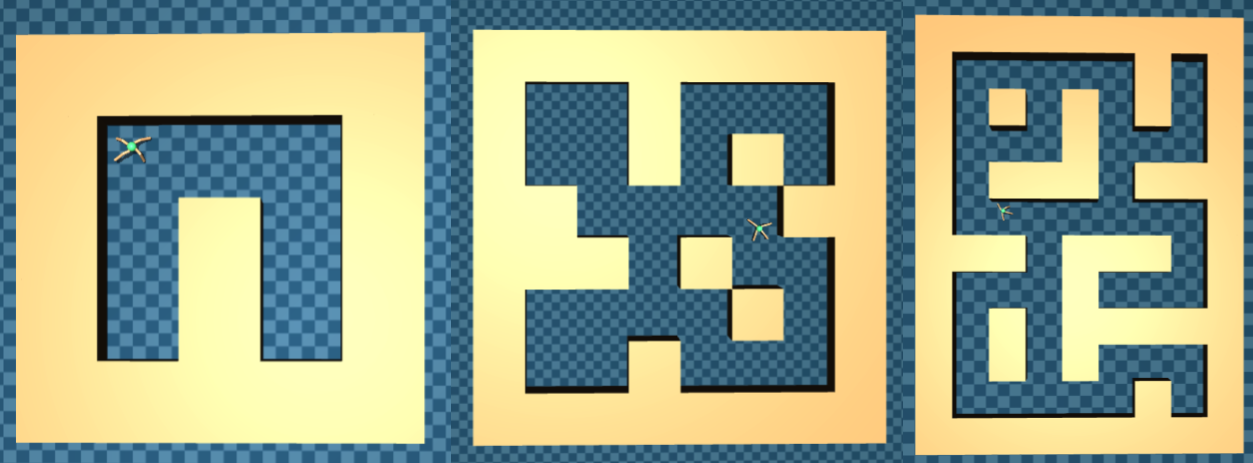}
    \caption{D4RL environment. Above: MuJoCo tasks of halfcheetah, hopper and walker2d. Below: AntMaze tasks with u-shape, medium, large maze.}
    \label{fig:env}
\end{figure}

\section{Experiments}
\label{sec:exp}
In this section, we provide empirical results to verify the effectiveness of our framework, aiming to answer the following four primary research questions (RQ): 

\textbf{RQ1.} How does SUNG compare with other state-of-the-arts when combined with different offline RL methods, across various types of tasks and datasets? 

\textbf{RQ2.} How do the key components of SUNG influence the overall offline-to-online performance? 

\textbf{RQ3.} How should we set hyper-parameters for SUNG in practice, and how challenging is this task?

\textbf{RQ4.} Can SUNG seamlessly integrate with other advanced RL techniques?

\subsection{Experimental Setup}
\subsubsection{Environment and Dataset}
We evaluate on the D4RL benchmark \cite{d4rl}, which provides various continuous-control tasks and datasets. We focus on MuJoCo and AntMaze domains, as shown in Fig~\ref{fig:env}. For MuJoCo tasks, we follow prior work \cite{pex, oema} to omit the evaluation on medium-expert and expert datasets, where offline pre-trained agents can already achieve expert-level performance and do not need further finetuning. Thus, we only focus on three kinds of quality: random (r), medium (m), and medium-replay (m-r). For AntMaze tasks, we select all the six settings, including u-shape, medium and large maze. For fair comparison, all the evaluations use datasets of the latest released "-v2" version that has fixed some potential bugs in the offline dataset. 

\subsubsection{Evaluation Protocol} 
We first perform 1M gradient steps for offline pretraining, and then perform 100K environment steps for online finetuning. While some prior works \cite{iql,pex,apl} take 1M environment steps for finetuning, we argue that 1M environment steps are even enough for an online RL agent to achieve expert-level performance. Thus, we believe finetuning for 100K environment steps is a more reasonable setting, which is also adopted by \cite{mcq,apl}. All the experiments are repeated with 5 different random seeds. For each evaluation, we run 10 episodes for MuJoCo tasks and 100 episodes for AntMaze tasks. We employ the normalized score, i.e., D4RL score, as a metric to assess the agent's performance. In this context, a score of 0 signifies random performance, while a score of 100 indicates expert-level performance. We report the average D4RL score over last 3 evaluations for MuJoCo tasks, and last 10 evaluations for AntMaze tasks.

\subsubsection{Backbone Offline RL Methods} To evaluate the generality of SUNG, we choose two representative offline RL methods as the backbone for pretraining, i.e., TD3+BC \cite{td3+bc} and CQL \cite{cql}. Concretely, TD3+BC extends TD3 with behavior cloning based policy constraint, while CQL extends SAC with value regularization. For AntMaze domains, we substitute TD3+BC with SPOT \cite{spot} due to its inferior performance. SPOT is another extension of TD3 but with a VAE-based density-guided policy constraint.

\begin{table}[t]
\centering
\caption{Hyper-parameters of online finetuning in SUNG.}
\begin{tabular}{@{}l|l|c|c|c@{}}
\toprule
 &  & TD3+BC & CQL & CQL-10 \\ \midrule
\multirow{4}{*}{SUNG} & Candidate action num $N$ & 100 & 100 & 20 \\
 & Finalist action num $k$ & 10 & 20 & 4 \\
 & Softmax temperature $\alpha$ & \multicolumn{3}{c}{1.0} \\
 & OOD sample percentage $p$ & 5 & 10 & 5 \\ \midrule
\multirow{3}{*}{OORB} & Online buffer size & \multicolumn{3}{c}{2e4} \\
 & Offline buffer size & \multicolumn{3}{c}{2e6} \\
 & OORB probability   $p_\mathrm{OORB}$ & \multicolumn{3}{c}{0.1} \\ \bottomrule
\end{tabular}
\label{tab:hyper-sung}
\end{table}

\subsubsection{Implementation details and hyper-parameters}
For VAE training, we follow the official implementation of BCQ \cite{bcq} and SPOT \cite{spot}. Unlike these methods, which use VAEs to estimate the behavior policy's action distribution, SUNG employs a VAE to model the state-action visitation density. Thus, we set the latent dimension of VAE to be twice the sum of the state and action dimensions.
Note that we simply take the hyper-parameters reported in SPOT \cite{spot} and omit hyper-parameter tuning for VAE training. Accordingly, further performance improvement is expected from careful tuning.

For online finetuning, we follow the same standard architecture and the same hyper-parameter of TD3+BC, CQL, and SPOT for all the baselines. Note that we do not tune any hyper-parameter of backbone RL and architecture in SUNG for fair comparison. We present the detailed hyper-parameter setting of SUNG in Table~\ref{tab:hyper-sung}.

In terms of baseline reproduction, we strictly follow their official implementation and hyper-parameters reported in their original paper to finetune agents pre-trained with TD3+BC, CQL and SPOT for MuJoCo and AntMaze domains. One exception is that the heavy ensemble technique is removed in BR for fair comparison. For fair comparison, we compare SUNG with the full implementation of \cite{br} as BRPQ in Section \ref{exp:ensemble}. Note that we also equip the baseline with OORB to achieved improved performance for fair comparison.

\begin{table*}[tbp]
\caption{Comparison of the averaged D4RL score on MuJoCo tasks with \textbf{TD3+BC} as the offline RL backbone method. We report the mean and standard deviation over 5 seeds.}
\begin{center}
\resizebox{\linewidth}{!}{
\begin{tabular}{@{}l|crrrrrrrr@{}}
\toprule \midrule
 & \multicolumn{1}{c}{\textbf{TD3+BC}} & \multicolumn{1}{c}{\textbf{offline-ft}} & \multicolumn{1}{c}{\textbf{online-ft}} & \multicolumn{1}{c}{\textbf{BR}} & \multicolumn{1}{c}{\textbf{O3F}} & \multicolumn{1}{c}{\textbf{APL}} & \multicolumn{1}{c}{\textbf{PEX}} & \multicolumn{1}{c}{\textbf{PROTO}} & \multicolumn{1}{c}{\textbf{SUNG}} \\ \midrule
\textbf{halfcheetah-r-v2} & 11.5 & 34.3\scalebox{0.8}{$\pm$2.4} & 57.7\scalebox{0.8}{$\pm$1.6} & 67.6\scalebox{0.8}{$\pm$11.9} & 71.4\scalebox{0.8}{$\pm$3.3} & 70.0\scalebox{0.8}{$\pm$4.7} & 53.4\scalebox{0.8}{$\pm$9.1} & 50.7\scalebox{0.8}{$\pm$1.3} & \textbf{76.6\scalebox{0.8}{$\pm$2.0}} \\
\textbf{hopper-r-v2} & 8.7 & 8.2\scalebox{0.8}{$\pm$0.2} & 11.2\scalebox{0.8}{$\pm$1.9} & 25.7\scalebox{0.8}{$\pm$8.7} & 11.7\scalebox{0.8}{$\pm$2.0} & 27.1\scalebox{0.8}{$\pm$14.0} & 37.4\scalebox{0.8}{$\pm$10.6} & 13.4\scalebox{0.8}{$\pm$9.5} & \textbf{38.7\scalebox{0.8}{$\pm$15.0}} \\
\textbf{walker2d-r-v2} & 5.4 & 7.0\scalebox{0.8}{$\pm$4.1} & 6.2\scalebox{0.8}{$\pm$3.9} & 9.9\scalebox{0.8}{$\pm$4.6} & 11.6\scalebox{0.8}{$\pm$5.1} & 13.8\scalebox{0.8}{$\pm$4.0} & \textbf{33.1\scalebox{0.8}{$\pm$16.4}} & 3.6\scalebox{0.8}{$\pm$3.1} & 14.1\scalebox{0.8}{$\pm$5.1} \\ \midrule
\textbf{halfcheetah-m-v2} & 48.0 & 49.3\scalebox{0.8}{$\pm$0.4} & 67.6\scalebox{0.8}{$\pm$2.4} & 79.5\scalebox{0.8}{$\pm$7.4} & 77.8\scalebox{0.8}{$\pm$1.1} & \textbf{80.9\scalebox{0.8}{$\pm$2.0}} & 52.3\scalebox{0.8}{$\pm$21.1} & 67.4\scalebox{0.8}{$\pm$1.8} & 80.7\scalebox{0.8}{$\pm$2.5} \\
\textbf{hopper-m-v2} & 61.5 & 58.8\scalebox{0.8}{$\pm$3.9} & 78.3\scalebox{0.8}{$\pm$32.7} & 93.1\scalebox{0.8}{$\pm$10.8} & \textbf{102.0\scalebox{0.8}{$\pm$2.0}} & 76.9\scalebox{0.8}{$\pm$24.2} & 73.9\scalebox{0.8}{$\pm$17.8} & 60.5\scalebox{0.8}{$\pm$23.3} & 101.8\scalebox{0.8}{$\pm$6.0} \\
\textbf{walker2d-m-v2} & 82.2 & 84.6\scalebox{0.8}{$\pm$1.2} & 68.2\scalebox{0.8}{$\pm$11.5} & 70.1\scalebox{0.8}{$\pm$21.4} & 97.1\scalebox{0.8}{$\pm$2.2} & 98.2\scalebox{0.8}{$\pm$13.5} & 56.7\scalebox{0.8}{$\pm$28.5} & 79.5\scalebox{0.8}{$\pm$9.4} & \textbf{113.5\scalebox{0.8}{$\pm$1.9}} \\ \midrule
\textbf{halfcheetah-m-r-v2} & 44.6 & 47.0\scalebox{0.8}{$\pm$0.9} & 66.3\scalebox{0.8}{$\pm$1.0} & 65.0\scalebox{0.8}{$\pm$14.6} & 67.6\scalebox{0.8}{$\pm$2.9} & \textbf{71.5\scalebox{0.8}{$\pm$1.3}} & 53.2\scalebox{0.8}{$\pm$9.8} & 61.0\scalebox{0.8}{$\pm$1.7} & 69.7\scalebox{0.8}{$\pm$3.4} \\
\textbf{hopper-m-r-v2} & 55.9 & 85.4\scalebox{0.8}{$\pm$7.6} & 89.9\scalebox{0.8}{$\pm$13.5} & 97.2\scalebox{0.8}{$\pm$13.9} & 97.6\scalebox{0.8}{$\pm$4.9} & 100.6\scalebox{0.8}{$\pm$9.8} & 90.8\scalebox{0.8}{$\pm$18.7} & 100.4\scalebox{0.8}{$\pm$1.0} & \textbf{101.3\scalebox{0.8}{$\pm$7.0}} \\
\textbf{walker2d-m-r-v2} & 71.7 & 80.2\scalebox{0.8}{$\pm$8.7} & 87.4\scalebox{0.8}{$\pm$4.0} & 82.7\scalebox{0.8}{$\pm$17.8} & 100.9\scalebox{0.8}{$\pm$3.7} & 108.2\scalebox{0.8}{$\pm$3.6} & 70.6\scalebox{0.8}{$\pm$13.3} & 93.7\scalebox{0.8}{$\pm$3.3} & \textbf{109.2\scalebox{0.8}{$\pm$1.9}} \\ \midrule
\textbf{Total} & \multicolumn{1}{c}{389.6} & \multicolumn{1}{c}{454.9} & \multicolumn{1}{c}{532.9} & \multicolumn{1}{c}{590.9} & \multicolumn{1}{c}{637.7} & \multicolumn{1}{c}{647.1} & \multicolumn{1}{c}{521.2} & \multicolumn{1}{c}{530.2} & \multicolumn{1}{c}{\textbf{705.7}} \\ \midrule

\textbf{Avg Improvement} & \multicolumn{1}{c}{/} & \multicolumn{1}{c}{16.76\%} & \multicolumn{1}{c}{36.78\%} & \multicolumn{1}{c}{51.66\%} & \multicolumn{1}{c}{63.68\%} & \multicolumn{1}{c}{66.09\%} & \multicolumn{1}{c}{33.77\%} & \multicolumn{1}{c}{36.08\%} & \multicolumn{1}{c}{\textbf{81.13\%}} \\

\midrule \bottomrule 
\end{tabular}
}
\end{center}
\label{tab:td3-mujoco}
\end{table*}

\begin{table*}[tbp]
\caption{Comparison of the averaged D4RL score on MuJoCo tasks with \textbf{CQL} as the offline RL backbone method. We report the mean and standard deviation over 5 seeds. Note that O3F is omitted due to its divergence in this setting.}
\begin{center}
\resizebox{0.9\linewidth}{!}{
\begin{tabular}{@{}l|crrrrrrr@{}}
\toprule \midrule 
 & \textbf{\textbf{CQL}} & \multicolumn{1}{c}{\textbf{offline-ft}} & \multicolumn{1}{c}{\textbf{online-ft}} & \multicolumn{1}{c}{\textbf{BR}} & \multicolumn{1}{c}{\textbf{APL}} & \multicolumn{1}{c}{\textbf{PEX}} & \multicolumn{1}{c}{\textbf{PROTO}} & \multicolumn{1}{c}{\textbf{SUNG}} \\ \midrule
\textbf{halfcheetah-r-v2} & \multicolumn{1}{c}{23.5} & 28.8\scalebox{0.8}{$\pm$2.3} & 50.2\scalebox{0.8}{$\pm$1.5} & 61.8\scalebox{0.8}{$\pm$12.3} & 67.7\scalebox{0.8}{$\pm$9.6} & 50.6\scalebox{0.8}{$\pm$2.2} & 24.9\scalebox{0.8}{$\pm$7.9} & \textbf{69.1\scalebox{0.8}{$\pm$9.2}} \\
\textbf{hopper-r-v2} & \multicolumn{1}{c}{6.4} & 31.2\scalebox{0.8}{$\pm$0.5} & 28.3\scalebox{0.8}{$\pm$6.2} & 23.8\scalebox{0.8}{$\pm$7.9} & 41.8\scalebox{0.8}{$\pm$22.0} & 34.3\scalebox{0.8}{$\pm$8.9} & 30.1\scalebox{0.8}{$\pm$3.1} & \textbf{44.3\scalebox{0.8}{$\pm$11.7}} \\
\textbf{walker2d-r-v2} & \multicolumn{1}{c}{4.5} & 5.6\scalebox{0.8}{$\pm$3.4} & 8.2\scalebox{0.8}{$\pm$5.6} & 4.0\scalebox{0.8}{$\pm$2.3} & 6.3\scalebox{0.8}{$\pm$1.8} & 10.7\scalebox{0.8}{$\pm$2.8} & 1.6\scalebox{0.8}{$\pm$2.3} & \textbf{14.5\scalebox{0.8}{$\pm$6.1}} \\ \midrule
\textbf{halfcheetah-m-v2} & \multicolumn{1}{c}{48.1} & 48.9\scalebox{0.8}{$\pm$0.2} & 52.1\scalebox{0.8}{$\pm$25.6} & 56.7\scalebox{0.8}{$\pm$28.5} & 44.7\scalebox{0.8}{$\pm$38.5} & 43.5\scalebox{0.8}{$\pm$2.4} & 52.1\scalebox{0.8}{$\pm$26.7} & \textbf{79.7\scalebox{0.8}{$\pm$1.0}} \\
\textbf{hopper-m-v2} & \multicolumn{1}{c}{73.7} & 74.1\scalebox{0.8}{$\pm$1.4} & 91.0\scalebox{0.8}{$\pm$10.4} & 97.7\scalebox{0.8}{$\pm$3.7} & 102.7\scalebox{0.8}{$\pm$3.1} & 46.3\scalebox{0.8}{$\pm$15.1} & 98.8\scalebox{0.8}{$\pm$5.1} & \textbf{104.1\scalebox{0.8}{$\pm$1.3}} \\
\textbf{walker2d-m-v2} & \multicolumn{1}{c}{84.3} & 83.5\scalebox{0.8}{$\pm$0.7} & 85.6\scalebox{0.8}{$\pm$7.6} & 81.7\scalebox{0.8}{$\pm$14.0} & 75.3\scalebox{0.8}{$\pm$25.7} & 34.0\scalebox{0.8}{$\pm$17.3} & 78.9\scalebox{0.8}{$\pm$11.2} & \textbf{86.0\scalebox{0.8}{$\pm$12.6}} \\ \midrule
\textbf{halfcheetah-m-r-v2} & \multicolumn{1}{c}{46.9} & 49.5\scalebox{0.8}{$\pm$0.3} & 61.3\scalebox{0.8}{$\pm$1.0} & 64.9\scalebox{0.8}{$\pm$5.8} & \textbf{78.6\scalebox{0.8}{$\pm$1.2}} & 45.5\scalebox{0.8}{$\pm$1.7} & 62.1\scalebox{0.8}{$\pm$4.2} & 75.6\scalebox{0.8}{$\pm$1.9} \\
\textbf{hopper-m-r-v2} & \multicolumn{1}{c}{96.0} & 95.0\scalebox{0.8}{$\pm$1.0} & 92.8\scalebox{0.8}{$\pm$19.5} & 88.5\scalebox{0.8}{$\pm$21.8} & 97.4\scalebox{0.8}{$\pm$9.5} & 66.5\scalebox{0.8}{$\pm$24.2} & 92.6\scalebox{0.8}{$\pm$19.4} & \textbf{101.9\scalebox{0.8}{$\pm$9.1}} \\
\textbf{walker2d-m-r-v2} & \multicolumn{1}{c}{83.4} & 84.5\scalebox{0.8}{$\pm$1.0} & 86.9\scalebox{0.8}{$\pm$12.4} & 78.8\scalebox{0.8}{$\pm$28.0} & 103.2\scalebox{0.8}{$\pm$19.0} & 40.1\scalebox{0.8}{$\pm$17.9} & 94.8\scalebox{0.8}{$\pm$1.8} & \textbf{108.2\scalebox{0.8}{$\pm$4.2}} \\ \midrule 
\textbf{Total} & 466.9 & \multicolumn{1}{c}{501.1} & \multicolumn{1}{c}{556.4} & \multicolumn{1}{c}{558.0} & \multicolumn{1}{c}{617.8} & \multicolumn{1}{c}{371.5} & \multicolumn{1}{c}{536.1} & \multicolumn{1}{c}{\textbf{683.4}} \\ \midrule

\textbf{Avg Improvement} & \multicolumn{1}{c}{/} & \multicolumn{1}{c}{7.32\%} & \multicolumn{1}{c}{19.16\%} & \multicolumn{1}{c}{19.51\%} & \multicolumn{1}{c}{32.31\%} & \multicolumn{1}{c}{-20.43\%} & \multicolumn{1}{c}{14.82\%} & \multicolumn{1}{c}{\textbf{46.36\%}} \\
\midrule \bottomrule
\end{tabular}
}
\end{center}
\label{tab:sac-mujoco}
\end{table*}

\subsubsection{Baselines} We compare SUNG with the following baselines for online finetuning: 
\begin{itemize}
    \item \textbf{offline-ft} leverages online interactions by performing offline RL objectives.
    \item \textbf{online-ft} leverages online interactions by performing online RL objectives that correspond to the ones used for pretraining. 
    \item \textbf{BR} \cite{br} prioritizes near-on-policy transitions from the replay buffer. 
    \item \textbf{O3F} \cite{o3f} utilizes knowledge contained in value functions to guide exploration. 
    \item \textbf{APL} \cite{apl} performs adaptive policy learning by incorporating different advantages of offline and online data. 
    \item \textbf{PEX} \cite{pex} trains a new policy from scratch by utilizing the pre-trained policy for exploration. 
    \item \textbf{PROTO} \cite{proto} introduces an iterative policy regularization scheme to achieve stable finetuning performance.
\end{itemize}
Note that we also include IQL \cite{iql}, ODT \cite{odt}, and ACA \cite{aca} for comparison with a separate subsection in Section \ref{section:odt}.

\subsection{Comparisons on D4RL Benchmarks (RQ1)}
In this subsection, we perform the overall performance comparison with different offline-to-online RL methods on both MuJoCo and AntMaze domains in D4RL benchmark. The full learning curves for all the settings are provided in the supplementary material.
\subsubsection{Results on MuJoCo} 
Results for MuJoCo domains with TD3+BC and CQL as backbone offline RL method are shown in Table \ref{tab:td3-mujoco} and \ref{tab:sac-mujoco}, respectively. We find that additional online interactions can bring further performance improvement in most settings. However, two exceptions are O3F and PEX, when combined with CQL. As pointed in \cite{o3f}, O3F is not compatible with value regularization based offline RL methods. In contrast, PEX trains a new policy from scratch, and cannot fully benefit from reusing good initialization from pretraining, thereby leading to limited performance.

Moreover, we remark that SUNG substantially outperforms previous state-of-the-art methods, exhibiting an additional 15.04\% and 14.05\% offline-to-online improvement over the best-performing baseline when combined with TD3+BC and CQL, respectively. This demonstrates the necessity of proper estimation and utilization of uncertainty for tackling both constrained exploratory behavior and state-action distribution shift. Moreover, this also verifies the generality of SUNG in facilitating online improvement for agents pretrained with different offline RL methods.

\begin{table*}[t]
\caption{Comparison of the averaged D4RL score on AntMaze tasks with \textbf{SPOT} and \textbf{CQL} as the offline RL backbone method. We report the mean and standard deviation over 5 seeds.}
\begin{tabular}{@{}l|ccrrr|ccrrr}
\toprule \midrule
\textbf{} & \textbf{SPOT} & \textbf{offline-ft} & \multicolumn{1}{c}{\textbf{O3F}} & \multicolumn{1}{c}{\textbf{APL}} & \multicolumn{1}{c|}{\textbf{SUNG}} & \textbf{CQL} & \textbf{offline-ft} & \multicolumn{1}{c}{\textbf{O3F}} & \multicolumn{1}{c}{\textbf{APL}} & \multicolumn{1}{c}{\textbf{SUNG}} \\ \midrule
\textbf{antmaze-umaze-v2} & 93.2 & 96.6\scalebox{0.8}{$\pm$0.4} & 96.9\scalebox{0.8}{$\pm$0.6} & \textbf{97.6\scalebox{0.8}{$\pm$1.0}} & 97.0\scalebox{0.8}{$\pm$0.8} & 88.6 & 87.9\scalebox{0.8}{$\pm$3.3} & \textbf{98.6\scalebox{0.8}{$\pm$0.3}} & 96.0\scalebox{0.8}{$\pm$1.3} & 96.9\scalebox{0.8}{$\pm$0.7} \\
\textbf{antmaze-umaze-diverse-v2} & 41.6 & 22.0\scalebox{0.8}{$\pm$18.1} & 65.3\scalebox{0.8}{$\pm$4.4} & 50.5\scalebox{0.8}{$\pm$30.7} & \textbf{71.7\scalebox{0.8}{$\pm$3.6}} & 41.0 & 45.9\scalebox{0.8}{$\pm$4.3} & 0.0\scalebox{0.8}{$\pm$0.0} & 0.0\scalebox{0.8}{$\pm$0.0} & \textbf{50.5\scalebox{0.8}{$\pm$9.6}} \\ \midrule
\textbf{antmaze-medium-play-v2} & 75.2 & 83.9\scalebox{0.8}{$\pm$2.2} & 85.6\scalebox{0.8}{$\pm$2.4} & 86.0\scalebox{0.8}{$\pm$2.3} & \textbf{88.6\scalebox{0.8}{$\pm$2.9}} & 63.8 & 75.4\scalebox{0.8}{$\pm$3.5} & 67.3\scalebox{0.8}{$\pm$17.0} & 22.8\scalebox{0.8}{$\pm$28.2} & \textbf{86.3\scalebox{0.8}{$\pm$2.1}} \\
\textbf{antmaze-medium-diverse-v2} & 73.0 & 84.6\scalebox{0.8}{$\pm$1.1} & 80.1\scalebox{0.8}{$\pm$7.8} & 86.0\scalebox{0.8}{$\pm$4.4} & \textbf{91.7\scalebox{0.8}{$\pm$1.6}} & 61.8 & 74.1\scalebox{0.8}{$\pm$2.0} & 80.8\scalebox{0.8}{$\pm$4.9} & 36.8\scalebox{0.8}{$\pm$33.4} & \textbf{85.6\scalebox{0.8}{$\pm$4.5}} \\ \midrule
\textbf{antmaze-large-play-v2} & 40.8 & 40.9\scalebox{0.8}{$\pm$4.6} & 38.8\scalebox{0.8}{$\pm$11.2} & 38.9\scalebox{0.8}{$\pm$4.7} & \textbf{45.7\scalebox{0.8}{$\pm$4.5}} & 32.0 & 41.8\scalebox{0.8}{$\pm$8.0} & 0.0\scalebox{0.8}{$\pm$0.0} & 0.0\scalebox{0.8}{$\pm$0.0} & \textbf{52.7\scalebox{0.8}{$\pm$9.8}} \\
\textbf{antmaze-large-diverse-v2} & 44.0 & 16.6\scalebox{0.8}{$\pm$2.2} & 3.0\scalebox{0.8}{$\pm$0.8} & 3.8\scalebox{0.8}{$\pm$1.6} & \textbf{19.8\scalebox{0.8}{$\pm$15.8}} & 32.6 & 35.3\scalebox{0.8}{$\pm$10.7} & 0.0\scalebox{0.8}{$\pm$0.0} & 0.0\scalebox{0.8}{$\pm$0.0} & \textbf{44.1\scalebox{0.8}{$\pm$12.3}} \\ \midrule
\textbf{Total} & 367.8 & 344.6 & \multicolumn{1}{c}{369.7} & \multicolumn{1}{c}{362.8} & \multicolumn{1}{c|}{\textbf{414.5}} & 319.6 & 360.4 & \multicolumn{1}{c}{246.7} & \multicolumn{1}{c}{155.6} & \multicolumn{1}{c}{\textbf{416.1}} \\ \midrule
\textbf{Avg Improvement} & / & -6.30\% & \multicolumn{1}{c}{0.51\%} & \multicolumn{1}{c}{-1.35\%} & \multicolumn{1}{c|}{\textbf{12.69\%}} & / & 12.76\% & \multicolumn{1}{c}{-22.80\%} & \multicolumn{1}{c}{-51.31\%} & \multicolumn{1}{c}{\textbf{30.19\%}} \\
\midrule \bottomrule
\end{tabular}
\label{tab:spot-antmaze}
\end{table*}

\subsubsection{Results on AntMaze}
We conduct experiments on more challenging AntMaze domains, which require the agents to deal with the sparse rewards and stitch fragments of sub-optimal trajectories. Results for AntMaze domains with SPOT and CQL as backbone offline RL method are shown in Table \ref{tab:spot-antmaze}. We omit the results of online-ft, BR, PEX, and PROTO due to their poor performance. This stems from the need for explicit or implicit behavior cloning that is offered by offline RL objectives, which is required to handle the sparse rewards and stitch sub-optimal sub-trajectories. However, online-ft, BR, PEX, and PROTO are developed on top of purely online RL methods, thus failing in most of the tasks.

Furthermore, we highlight that SUNG exhibits an additional 12.43\% and 17.43\% offline-to-online improvement over the best-performing baseline when combined with SPOT and CQL, respectively. This demonstrates the superiority of SUNG for discovering informative state-action pairs that can lead to high outcomes, thereby providing valuable behavior patterns for agents to learn from. Moreover, SUNG achieves flexible trade-off between offline and online RL objectives by tuning $p$, which enables stitching sub-optimal sub-trajectories and overcoming the sparse reward issue. 

In the following subsections, we mainly focus on analyses in MuJoCo domains, given their wide adoption in previous works \cite{oema, br, apl}.

\subsection{Comparison with Other Offline-to-Online RL Methods (RQ1)}
\label{section:odt}
In this subsection, we compare SUNG with other offline-to-online RL methods that are designed for one specific offline RL method. Concretely, we select the following baselines: IQL \cite{iql}, ODT \cite{odt} and ACA \cite{aca}. We follow the original experimental setting in ODT to take 200K environment steps and only consider tasks with medium and medium-replay offline datasets for online finetuning to achieve fair comparison. The results are directly copied from \cite{aca}.

\begin{figure}[t]
\centering
\includegraphics[width=\linewidth]{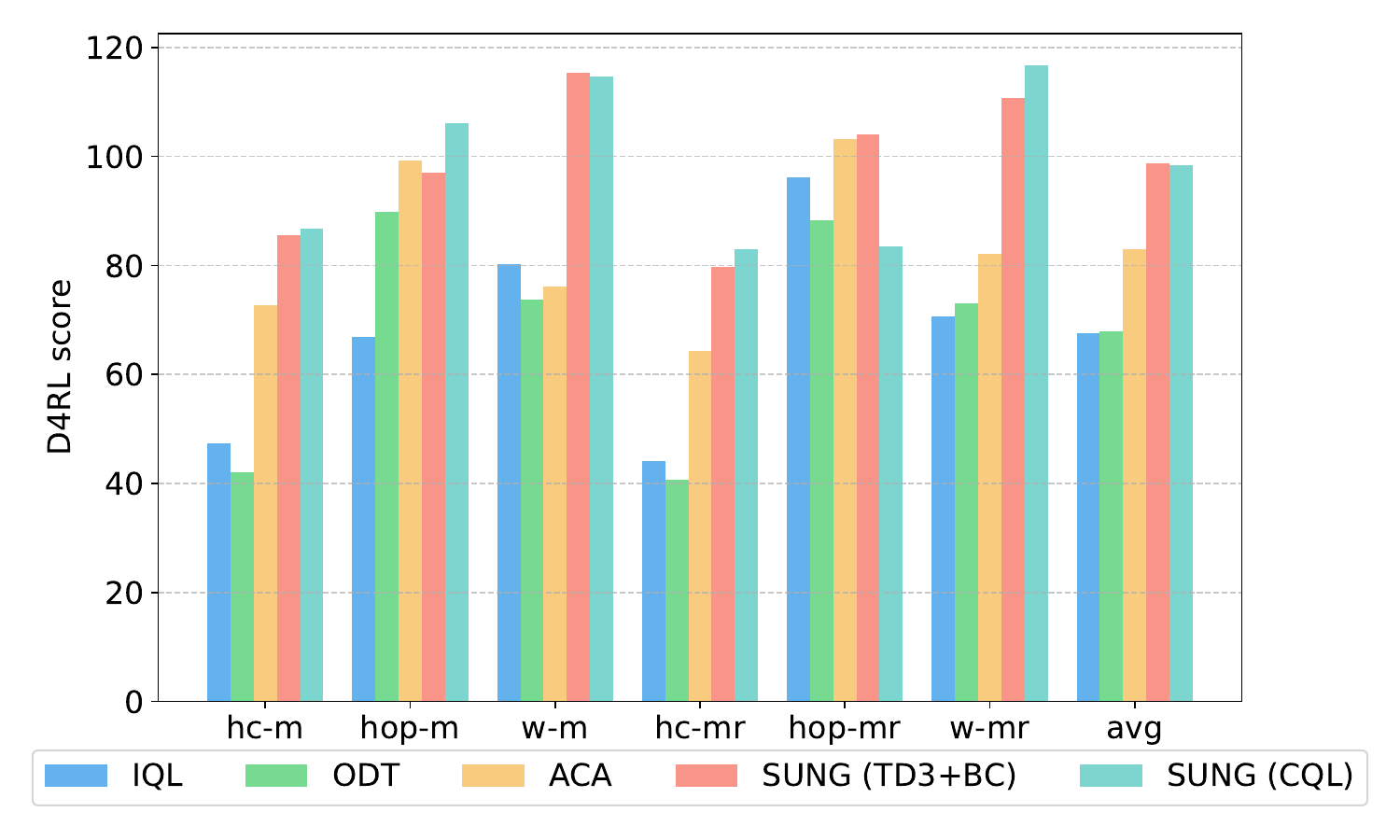}
\caption{Final performance of different offline-to-online RL methods with 200K environment steps. We report the mean D4RL score across 5 random seeds. hc = halfcheetah, hop = hopper, w = walker2d, m = medium, mr = medium-replay.}
\label{fig:others}
\end{figure}

We present experimental results of the final performance of different offline-to-online RL methods with 200K environment steps. As shown in Fig.~\ref{fig:others}, for almost all the settings, SUNG when combined with either TD3+BC or CQL outperforms other methods by a large margin in terms of final performance. IQL presents conservative online finetuning, which can stably improve the performance but derive limited performance under a limited interaction setting, i.e., 200K environment steps. ODT learns a reward-conditioned policy in a supervised learning paradigm, thereby struggling to fully benefit from the trail-and-error paradigm used in traditional online RL. However, ODT enables users to customize the behavior of the agents by specifying different conditions, which is not featured by traditional RL.

\subsection{Ablation Studies (RQ2)} 
\begin{figure*}
    \centering
    \includegraphics[width=\linewidth]{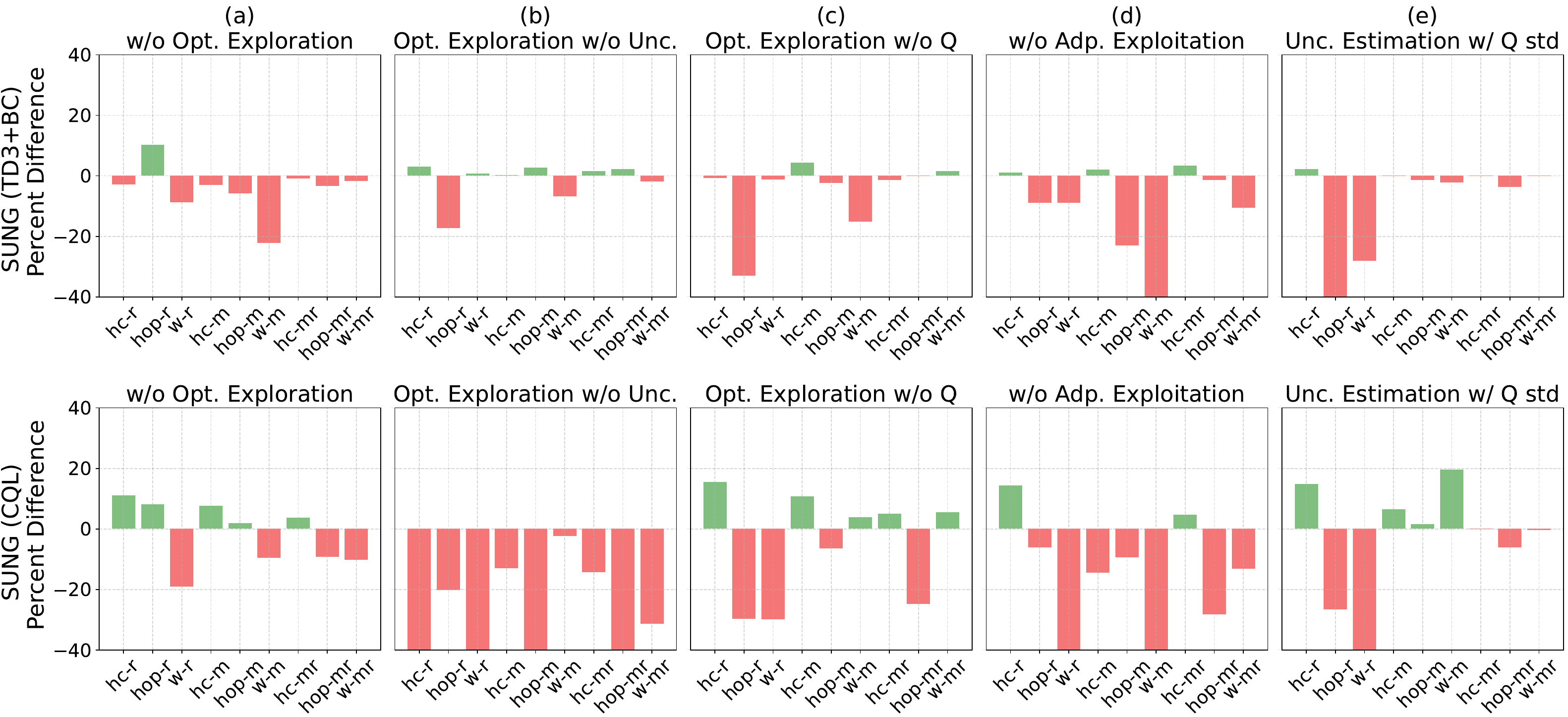}
    \caption{Performance difference of an ablation study of SUNG combined with TD3+BC and CQL, compared with the full algorithm. We report the mean performance difference across 5 different random seeds. Opt. = Optimistic, Unc. = Uncertainty, Adp. = Adaptive. hc = halfcheetah, hop = hopper, w = walker2d, r = random, m = medium, mr = medium-replay.}
    \label{fig:ablation}
\end{figure*}

In this subsection, we showcase experimental results in Fig. \ref{fig:ablation} to validate the effectiveness of each component of SUNG.
\subsubsection{Ablation on Optimistic Exploration}
We evaluate \textbf{(a) SUNG w/o Optimistic Exploration}, i.e., we replace the proposed optimistic exploration strategy by the default exploration provided by TD3 and SAC. For TD3, we select action with Gaussian noise for exploration, i.e., $\pi_{E}(s)= \pi_\phi(s)+\epsilon, \epsilon \sim \mathcal{N}(0,\delta)$. For SAC, we sample action from the policy for exploration, i.e., $\pi_{E}(s)=a, a\sim \pi_\phi(\cdot|s)$. We find that the removal brings significant negative effects when combined with TD3+BC, but no remarkable effects when combined with CQL. The reason for this discrepancy is that CQL is built on top of SAC, which can gradually recover exploration power through max-entropy RL framework during online finetuning. 

Moreover, we ablate two key components in the optimistic exploration strategy, uncertainty and Q value, to derive two variant strategies: \textbf{(b) Optimistic Exploration w/o Uncertainty} and \textbf{(c) Optimistic Exploration w/o Q}. For the former, we remove the uncertainty from the proposed exploration strategy to greedily select action that maximizes Q value, i.e., $\pi_{E}(s)=\arg\max_{a\in\mathcal{C}}Q(s,a)$. For the latter, we remove the Q value from the proposed exploration strategy to greedily select action that maximizes the uncertainty, i.e., $\pi_{E}(s)=\arg\max_{a\in\mathcal{C}}\mathcal{U}(s,a)$. As expected, both ablations degrade the performance to varying degrees, underscoring the significance of both Q value and uncertainty for exploration. Notably, the removal of uncertainty in SUNG when combined with CQL causes significant performance degradation, which can be attributed to similar reason detailed in \cite{o3f}.

\subsubsection{Ablation on Adaptive Exploitation}
We evaluate \textbf{(d) SUNG w/o Adaptive Exploitation}, which replaces the proposed adaptive exploitation method by the standard online RL objective of TD3 and SAC, respectively. From the corresponding experimental results, we note a significant degradation in performance for finetuning TD3+BC and CQL across most tasks. This emphasizes the necessity of addressing the state-action distribution shift issue.

\subsubsection{Ablation on Uncertainty Quantification} We introduce the variant: \textbf{(e) Uncertainty Estimation w/ Q std}, which replaces the VAE-based density estimator with the standard deviation of double Q values for uncertainty quantification, i.e., $\mathcal{U}(s,a)=\sigma(Q_{\theta_1}(s,a), Q_{\theta_2}(s,a))=|Q_{\theta_1}(s,a)-Q_{\theta_2}(s,a)|$. As expected, the results show that such replacement leads to performance deterioration on most tasks, since it is not sufficient to provide reliable uncertainty quantification. While utilizing the ensemble technique can eliminate this issue \cite{sunrise,ucb-rl}, it can significantly increase computational costs. We leave the exploration of alternative methods as future work.

\begin{figure}[t]
\centering
\includegraphics[width=0.8\linewidth]{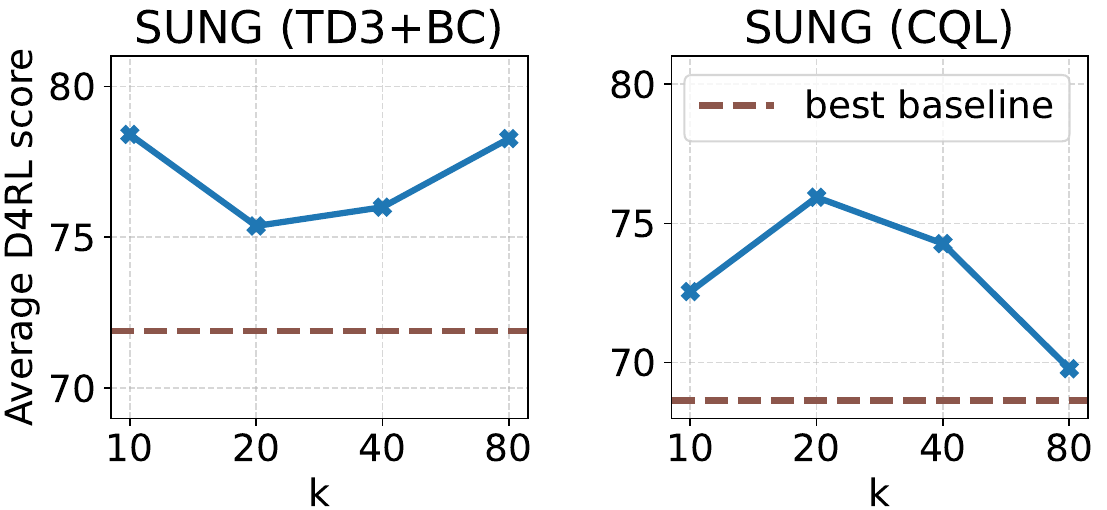}
\caption{Comparing performance on MuJoCo domains with different finalist action set size $k$ used in the optimistic exploration strategy.}
\label{fig:exploration}
\end{figure}

\begin{figure}[t]
\centering
\includegraphics[width=0.8\linewidth]{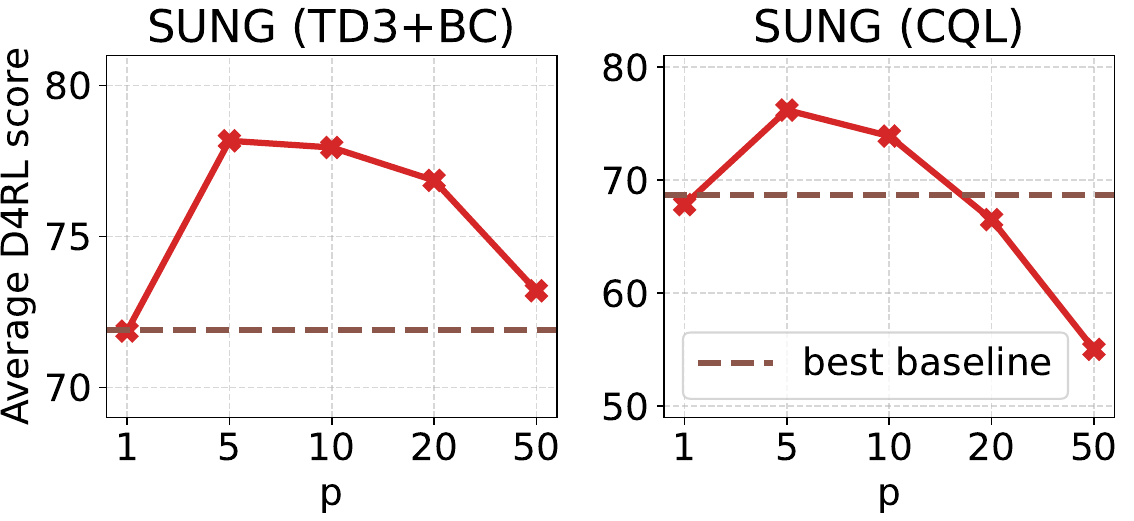}
\caption{Comparing performance on MuJoCo domains with varying percent $p$ of identified OOD samples from the mini-batch used in the adaptive exploitation method.}
\label{fig:exploitation}
\end{figure}

\subsection{Hyper-parameter Analysis (RQ3)}
In this subsection, we give a hyper-parameter analysis of SUNG, focusing on two important hyperparameters: final action set size $k$ and identified OOD sample percentage $p$.

\subsubsection{Analysis of Finalist Action Set Size $k$} First, we investigate the finalist action set size $k$ in optimistic exploration. We tune it in the range of $\{10,20,40,80\}$. The aggregated results averaged over the MuJoCo domains are shown in Fig. \ref{fig:exploration}. Note that we also report the performance of the best baseline. As expected, SUNG with TD3+BC can accommodate preference for both high uncertainty and high value, whereas SUNG with CQL can only accommodate preference for high uncertainty. Furthermore, we remark that SUNG consistently outperforms the best baseline with any choice of $k$, which further verifies the superiority of SUNG.

\subsubsection{Analysis of Adaptive Exploitation} Moreover, we explore the choice of percent $p$ of identified OOD samples from the mini-batch in adaptive exploitation. We tune it in the range of $\{1,5,10,20,50\}$. The aggregated results averaged over MuJoCo domains are shown in Fig. \ref{fig:exploitation}. Predictably, setting $p$ too conservatively or too aggressively may adversely degrade the performance. We find that $p=5$ performs the best when combined with either TD3+BC or CQL, achieving the desired trade-off between performance and stability. Furthermore, we want to highlight that SUNG outperforms the best baseline with most choices of $p$, which further emphasizes the superiority of SUNG.

\subsection{SUNG with Other Advanced Techniques (RQ4)}
\label{exp:ensemble}
\begin{table}[tbp]
\caption{Comparison of the averaged D4RL score on MuJoCo tasks with \textbf{CQL-10} as the offline RL backbone method. We report the mean and standard deviation over 5 seeds.}
\begin{center}
\resizebox{\linewidth}{!}{
\begin{tabular}{@{}l|crrrr@{}}
\toprule \midrule
\textbf{} & \textbf{CQL-10} & \multicolumn{1}{c}{\textbf{BRPQ}} & \multicolumn{1}{c}{\textbf{\begin{tabular}[c]{@{}c@{}}RLPD \\ (UTD=20)\end{tabular}}} & \multicolumn{1}{c}{\textbf{\begin{tabular}[c]{@{}c@{}}SUNG \\ (UTD=1)\end{tabular}}} & \multicolumn{1}{c}{\textbf{\begin{tabular}[c]{@{}c@{}}SUNG \\ (UTD=5)\end{tabular}}} \\ \midrule
\textbf{halfcheetah-r-v2} & 29.9 & 85.8\scalebox{0.8}{$\pm$17.2} & 72.5\scalebox{0.8}{$\pm$5.9} & \textbf{92.7\scalebox{0.8}{$\pm$4.2}} & \textbf{97.5\scalebox{0.8}{$\pm$5.0}} \\
\textbf{hopper-r-v2} & 7.4 & 28.4\scalebox{0.8}{$\pm$12.0} & 87.8\scalebox{0.8}{$\pm$14.0} & 62.5\scalebox{0.8}{$\pm$13.8} & \textbf{89.7\scalebox{0.8}{$\pm$28.4}} \\
\textbf{walker2d-r-v2} & 21.6 & 16.1\scalebox{0.8}{$\pm$4.0} & 65.7\scalebox{0.8}{$\pm$16.4} & 33.0\scalebox{0.8}{$\pm$19.6} & \textbf{73.1\scalebox{0.8}{$\pm$33.0}} \\ \midrule
\textbf{halfcheetah-m-v2} & 55.0 & 80.3\scalebox{0.8}{$\pm$20.5} & 84.2\scalebox{0.8}{$\pm$2.2} & \textbf{96.6\scalebox{0.8}{$\pm$1.0}} & \textbf{105.5\scalebox{0.8}{$\pm$3.6}} \\
\textbf{hopper-m-v2} & 66.9 & 79.5\scalebox{0.8}{$\pm$19.0} & 98.1\scalebox{0.8}{$\pm$12.2} & \textbf{111.4\scalebox{0.8}{$\pm$0.8}} & 86.3\scalebox{0.8}{$\pm$24.2} \\
\textbf{walker2d-m-v2} & 83.2 & 74.4\scalebox{0.8}{$\pm$13.9} & \textbf{114.3\scalebox{0.8}{$\pm$2.2}} & 113.8\scalebox{0.8}{$\pm$2.6} & 111.3\scalebox{0.8}{$\pm$25.3} \\ \midrule
\textbf{halfcheetah-m-r-v2} & 52.6 & 74.8\scalebox{0.8}{$\pm$20.7} & 80.3\scalebox{0.8}{$\pm$1.9} & \textbf{92.5\scalebox{0.8}{$\pm$0.5}} & \textbf{96.7\scalebox{0.8}{$\pm$6.1}} \\
\textbf{hopper-m-r-v2} & 102.3 & 71.6\scalebox{0.8}{$\pm$10.1} & 75.0\scalebox{0.8}{$\pm$17.9} & \textbf{100.7\scalebox{0.8}{$\pm$13.6}} & 64.9\scalebox{0.8}{$\pm$14.8} \\
\textbf{walker2d-m-r-v2} & 82.1 & 71.7\scalebox{0.8}{$\pm$11.2} & 108.5\scalebox{0.8}{$\pm$2.8} & 105.1\scalebox{0.8}{$\pm$14.6} & \textbf{117.3\scalebox{0.8}{$\pm$5.8}} \\ \midrule
\textbf{Total} & 501.0 & \multicolumn{1}{c}{582.6} & \multicolumn{1}{c}{786.4} & \multicolumn{1}{c}{\textbf{808.2}} & \multicolumn{1}{c}{\textbf{842.3}} \\ \midrule 
\textbf{Avg Improvement} & / & \multicolumn{1}{c}{16.28\%} & \multicolumn{1}{c}{56.96\%} & \multicolumn{1}{c}{\textbf{61.31\%}} & \multicolumn{1}{c}{\textbf{68.12\%}} \\ 
\midrule \bottomrule
\end{tabular}
}
\end{center}
\label{tab:cql-10}
\end{table}

In this subsection, we aim to illustrate that SUNG can be seamlessly compatible with other advanced RL techniques, leading to further improved finetuning performance. Particularly, we focus on: \textbf{(1) Ensemble Q learning}, as demonstrated in \cite{ensemble, br}, exhibits robust performance in offline pretraining and online finetuning. Nevertheless, it is crucial to acknowledge that the ensemble mechanism introduces additional computational costs. \textbf{(2) High update-to-data (UTD) ratio} \cite{utdr, onlineRLwithOfflineData}, as discussed in \cite{utdr, onlineRLwithOfflineData}, enhances sample efficiency by maximizing the number of gradient updates per environment step. Nonetheless, this efficiency comes at the expense of increased computational costs.

Specifically, we follow previous works \cite{br, onlineRLwithOfflineData} to utilize CQL-10, a variant of CQL employing ensemble Q learning with 10 Q functions, for offline pretraining. We compare SUNG against two baselines: \textbf{(1) BRPQ} \cite{br} introduces both prioritized replay buffer and pessimistic Q ensembles for offline-to-online RL. \textbf{(2) RLPD} \cite{onlineRLwithOfflineData} is a recent state-of-the-art method that utilizes offline data to accelerate online RL, benefiting from both the ensemble technique and high UTD ratio. For a fair comparison, we present results of SUNG with UTD ratio of both 1 and 5.

We present the experimental results in Table \ref{tab:cql-10}. We can observe that SUNG employing the ensemble Q learning technique, indicated by UTD=1, exhibits a significant performance advantage over BRPQ, showcasing an additional 45.03\% improvement in offline-to-online performance. In contrast, SUNG, when combined with both ensemble Q learning and a high UTD ratio, denoted as UTD=5, attains results that are either superior or, at the very least, comparable to those of RLPD, yielding an additional 11.16\% improvement in offline-to-online performance. The findings above underscore the seamless synergy of SUNG with other advanced RL techniques for enhanced finetuning performance.

Furthermore, we remark that SUNG with UTD ratio of 1 exhibits weak results for hopper-r and walker2d-r, suggesting that poorly initialized policies are challenging to recover through low UTD ratio online finetuning. In contrast, SUNG with UTD ratio of 5 demonstrates competitive results in these settings. However, we notice that SUNG with UTD ratio of 5 shows subpar performance in hopper-m and hopper-m-r, possibly due to the statistical overfitting \cite{utdr}. As suggested in previous works, regularization approaches, such as simple L2 normalization, layer normalization, dropout and random ensemble distillation, can eliminate this issue. We consider this avenue for future work.

\begin{table}[t]
\caption{Comparison of training time with \textbf{CQL} as the offline RL backbone method.}
\resizebox{\linewidth}{!}{
\begin{tabular}{@{}cccccccc@{}}
\toprule
\hline
\textbf{offline-ft} & \textbf{online-ft} & \textbf{BR} & \textbf{APL} & \textbf{PEX} & \textbf{PROTO} & \textbf{\begin{tabular}[c]{@{}c@{}}SUNG\\  w/ VAE\end{tabular}} & \textbf{\begin{tabular}[c]{@{}c@{}}SUNG \\ w/ Q-Ensemble\end{tabular}} \\ \midrule
1h & 1h & 0.8h & 1h & 0.6h & 0.8h & 0.7h & 2.4h \\ 
\hline
\bottomrule
\end{tabular}
}
\label{exp:time}
\end{table}

\subsection{Comparison on Training Time}
In this subsection, we compare the training time of various offline-to-online RL methods. The results are summarized in Table \ref{exp:time}, where we use CQL as the backbone method and report the time required for 100K environment steps in the halfcheetah-m-v2 setting. The experiments are conducted on a single NVIDIA RTX 3090 GPU and Intel Xeon Platinum 8255C CPU at 2.50GHz. Notably, we include SUNG w/ Q-Ensemble, which utilizes 10 Q-networks for uncertainty estimation. As shown in the results, all baseline methods require approximately 1 hour of training, except for SUNG w/ Q-Ensemble, which takes 2.4 hours. This also highlights the computational efficiency of our proposed VAE-based uncertainty estimation methods.

\section{Conclusion}
\label{sec:conclusion}
This paper investigates the effective finetuning methods of pretrained offline RL agents via online interactions. We introduce SUNG, a simple yet effective uncertainty-guided framework, designed to address the challenges of constrained exploratory behavior and state-action distribution shift. SUNG employs a VAE estimator to quantify state-action uncertainty. The optimistic exploration strategy, facilitated by a bi-level action selection mechanism, enables SUNG to efficiently explore the state-action space. Furthermore, SUNG incorporates an adaptive exploitation method with OOD sample identification, creating a seamless bridge between the offline and online learning stages. Extensive empirical results, spanning various backbone offline RL methods, environments, and datasets, confirm the superiority of SUNG. Notably, SUNG demonstrates robust finetuning performance across diverse hyperparameter settings and can be seamlessly integrated with different advanced RL techniques to achieve further improved finetuning results.

\section{Limitation and Future Work}
While achieving promising performance, the proposed framework also has some inherent limitations. SUNG investigates the offline-to-online RL setting to solve tasks whose inputs are the states of the simulated environment. It would be interesting to see whether SUNG is also superior with the visual RL setting. 
We also remark that the exploration of its application on real-world tasks is also an interesting future work.
Furthermore, SUNG may struggle in finetuning pretrained agents from offline datasets with low-quality behaviors, highlighting the need for more robust policy improvement methods within offline-to-online RL algorithms.
Finally, further investigation on uncertainty estimation, replay buffer, and other backbone offline RL methods needs to be explored. 

\bibliographystyle{IEEEtran}
\bibliography{IEEEabrv, references.bib}

\newpage
\appendices
\section{Further Discussion on Other Uncertainty Quantification Methods}
In this section, we provide more detailed discussion with other uncertainty quantification methods in RL, where we mainly categorize them as (1) Distributional RL, (2) Ensemble based methods, (3) Bootstrapping based methods, (4) Monte Carlo dropout based methods, and (5) Generative models based methods.

\begin{itemize}
  \item \textbf{Distributional RL}, such as C51 \cite{c51}, QR-DQN \cite{qr-dqn}, and IQN \cite{iqn}, aims to model the aleatoric uncertainty on return distribution that originates from inherent stochasticity of the environment and agent interactions in RL. Different from them, we aim to model epistemic uncertainty that originates from the insufficient knowledge of the environment to guide exploration and exploitation in offline-to-online RL. Thus, the investigated problems of them are inherently different, and the uncertainty quantification methods in distributional RL cannot be applied to general offline-to-online RL setting.

  \item \textbf{Ensemble based methods}, such as SUNRISE \cite{sunrise}, utilize the variance of the predicted Q values from the ensemble Q networks as the uncertainty. However, their applicability to the offline-to-online RL setting is limited, as they require the offline pretraining phase to also employ ensemble-based offline RL algorithms. In contrast, the VAE-based uncertainty quantification method in SUNG can be generally applicable. We also provide an empirical comparison in Appendix E.A to highlight the effectiveness of the VAE-based method.

  \item \textbf{Bootstrapping based methods}, such as Bootstrapped DQN \cite{bdqn}, estimate the diversity of multiple heads of Q networks that are trained from different sample subsets as uncertainty. Similar to ensemble based methods, they also require the offline pretraining phase to employ bootstrapping-based offline RL algorithms, which limit their applicability in the offline-to-online RL setting. We do not include the empirical comparison of this method due to its similarity to ensemble-based methods.

  \item \textbf{Monte Carlo dropout based methods}, such as UWAC \cite{uwac}, quantify epistemic uncertainty by the variance of Q values from multiple forward passes of the dropout-enabled Q network. Since modern offline RL algorithms typically do not utilize the dropout technique, the applicability of this method is also limited in the offline-to-online RL setting. We try to make comparison between the proposed VAE method and Monte Carlo dropout based methods; however, as modern offline RL algorithms typically utilize shallow neural networks (e.g., three-layer MLP in TD3+BC and CQL), the dropout technique makes the training process quite unstable in practice. We leave the exploration of stable Monte Carlo dropout based offline-to-online RL methods as future work.

  \item \textbf{Generative model based methods}, such as CPQ \cite{cpq}, estimate the action density of the behavior policy distribution of the offline dataset as the uncertainty to penalize offline RL polices in generating OOD actions. In contrast, we utilize VAE to estimate density of state-action pairs as uncertainty quantification, which can guide both exploration and exploitation in offline-to-online RL. Although diffusion models \cite{diffusion-1, diffusion-2} are regarded as stronger generative models than VAEs, they require higher computational costs for training and inference. Moreover, the hyper-parameter tuning for diffusion models is non-trivial. As such, we adopt VAE in this work and leave the extension to diffusion models as future work.
  
\end{itemize}

\begin{figure}
    \centering
    \includegraphics[width=\linewidth]{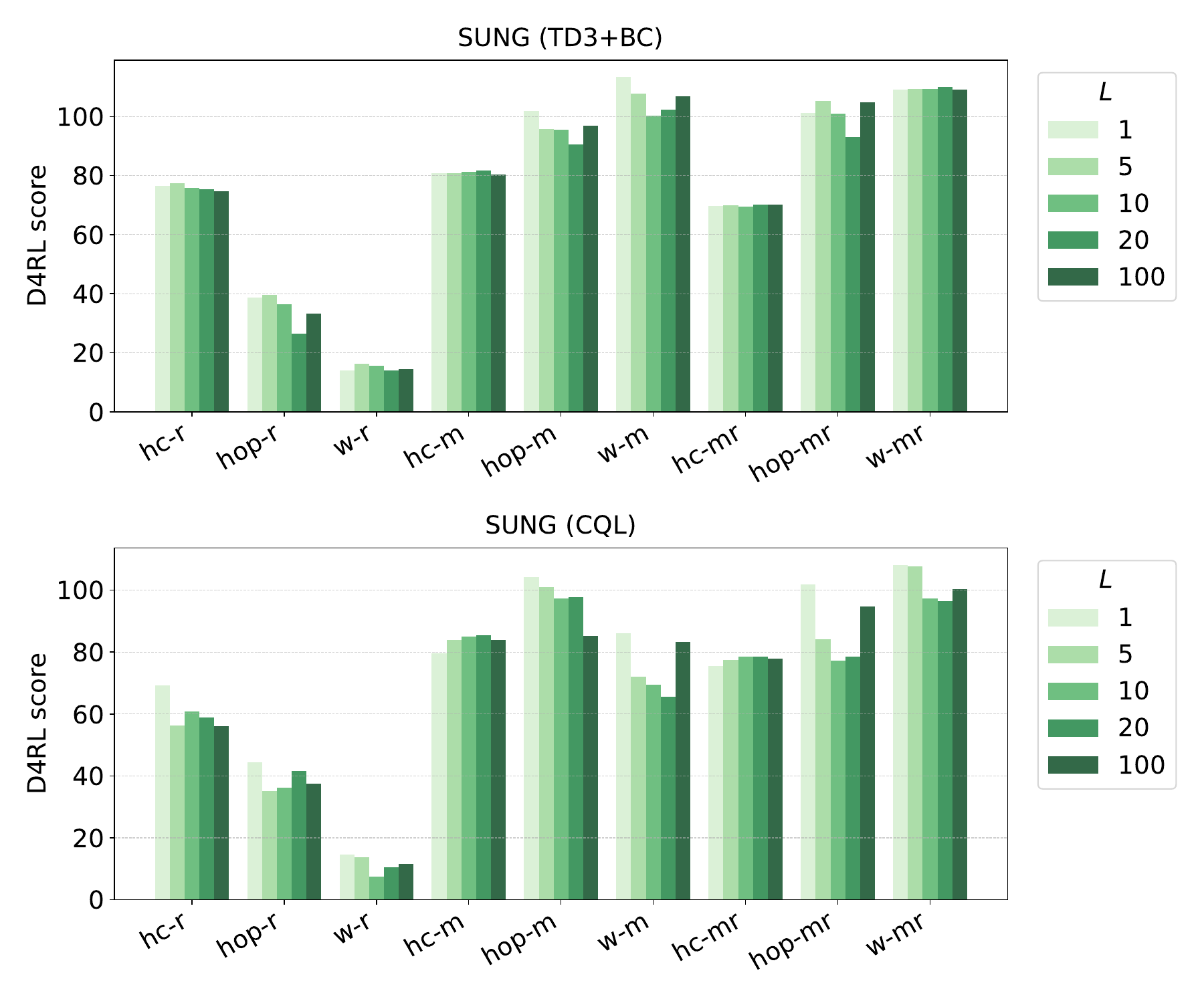}
    \caption{Hyper-parameter analysis on $L$ of SUNG combined with TD3+BC and CQL, where a larger $L$ means lower bias on uncertainty quantification.}
    \label{fig:tigher-bound}
\end{figure}

\section{Discussion on bias in Density Estimation}
As stated earlier, there theoretically exists a bias between ELBO and the negative log likelihood of the state-action density, since we have $-\log p_\psi(s,a|z)=\mathcal{L}_{\mathrm{ELBO}}- D_{\mathrm{KL}}\left[q_\varphi(z|s,a)||p(z)\right]$. To achieve further bias reduction, we follow \cite{iwae, spot} to utilize the importance sampling technique to derive the density estimation as:
\begin{equation}
\begin{aligned}
   \log p(s,a) &= \mathbb{E}_{q_\varphi(z|s,a)}\left[\log\frac{p_\psi(s,a,z)}{q_\varphi(z|s,a)}\right] \\
   &\approx \mathbb{E}_{q_\varphi(z^{(l)}|s,a)}\left[\log\frac{1}{L}\sum_{l=1}^L\frac{p_\psi(s,a,z^{(l)})}{q_\varphi(z^{(l)}|s,a)}\right].
\end{aligned}
\end{equation}
As theoretically shown in \cite{iwae}, this estimator is the lower bound of the log likelihood of the state-action density; the bound becomes tighter as $L$ increases and it converges to exact equality in the limit as $L\rightarrow\infty$. Particularly, when $L=1$, it is exactly the ELBO estimator. 

Now, we give empirical analyses on the hyper-parameter $L$ to demonstrate how the bias in uncertainty quantification impacts the performance of SUNG. We present results of $L\in[1,5,10,20,100]$ in Fig.~\ref{fig:tigher-bound}, where a larger $L$ means lower bias on uncertainty quantification. We can find that all the variants yield similar performance, thus the slight bias in uncertainty quantification would not significantly impact the performance of SUNG. As such, we set $L=1$ (i.e., ELBO estimator) in this paper, due to its simplicity and effectiveness.

\begin{figure*}
    \centering
    \includegraphics[width=\linewidth]{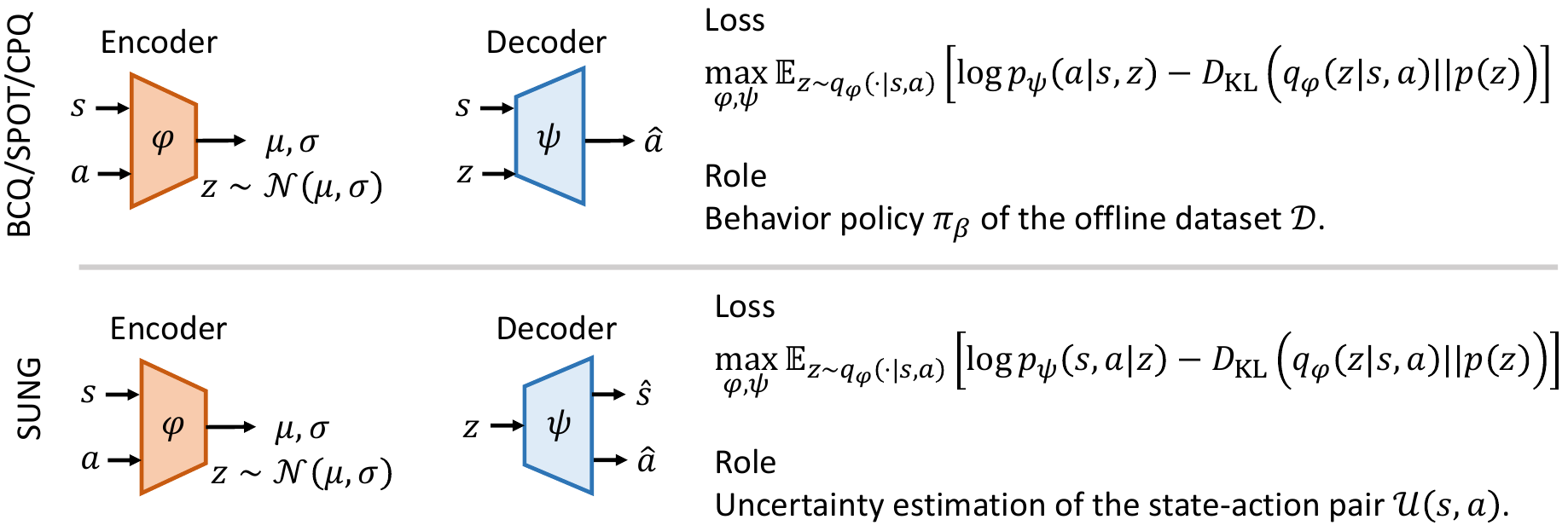}
    \caption{Comparison of VAE usage in offline RL and offline-to-online RL.}
    \label{fig:vae-diff}
\end{figure*}
\section{More Implementation Details of VAE}
\begin{figure}[h]
    \centering
    \includegraphics[width=\linewidth]{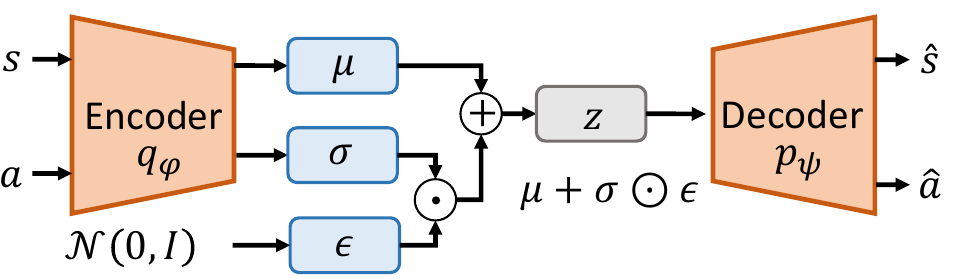}
    \caption{Illustration of the VAE architecture.}
    \label{fig:vae}
\end{figure}

We present more implementation details for the VAE here. As the VAE is trained to estimate the state-action density in our work, we design its architecture of as an encoder of a three-layer MLP ($d_{\mathcal{S}}+d_{\mathcal{A}}$, 750, 750, $2(d_{\mathcal{S}}+d_{\mathcal{A}})$) and a decoder of a three-layer MLP ($2(d_{\mathcal{S}}+d_{\mathcal{A}})$, 750, 750, $d_{\mathcal{S}}+d_{\mathcal{A}}$), where $d_{\mathcal{S}}$ and $d_{\mathcal{A}}$ denote the state dimension and action dimension. We illustrate the VAE architecture in Fig.~\ref{fig:vae}. During the training of VAE, the encoder encodes the state-action pair $(s,a)$ and outputs the mean vector $\mu$ and standard deviation $\sigma$, which specify the normal distribution $\mathcal{N}(\mu, \sigma^2)$ for the hidden representation $z$. Since the sampling process is non-differentiable, we apply the reparameterization trick to sample $z \sim \mathcal{N}(\mu, \sigma^2)$ by sampling a standard normal distribution $\epsilon \sim \mathcal{N}(0,I)$ and mapping to $\mathcal{N}(\mu, \sigma^2)$ as $z=\mu+\sigma\odot\epsilon$. Then, the hidden representation is decoded by the encoder as the generated state $\hat{s}$ and action $\hat{a}$. As such, the parameters of the VAE can be optimized by maximizing the ELBO. During the inference of VAE, we only perform the forward pass to derive the ELBO as the uncertainty estimation.

\section{Full Learning Curves}
\begin{figure*}[h]
    \centering
    \includegraphics[width=\linewidth]{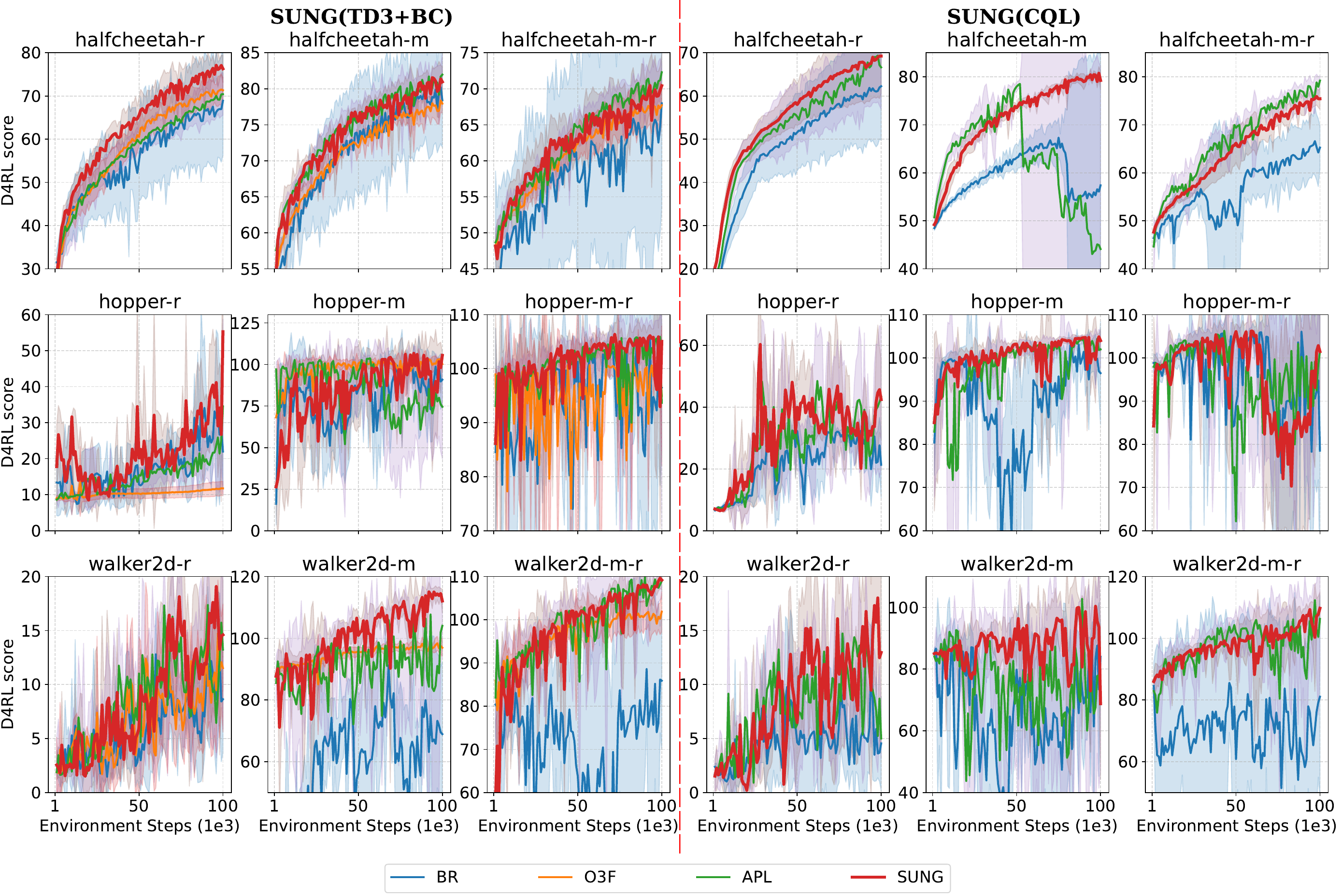}
    \caption{Learning curves of SUNG, when combined with TD3+BC and CQL, for 100K environment steps on \textbf{MuJoCo} tasks. We also include three best-performing baselines, BR, O3F and APL, for comparison. w = Walker, r = random, m = medium, m-r = medium-replay. We report the mean D4RL score and standard deviation over 5 training seeds with 10 evaluation episodes each. Note that O3F is omitted when combined with CQL due to its divergence.}
    \label{fig:plot}
\end{figure*}

\begin{figure*}[h]
    \centering
    \includegraphics[width=\linewidth]{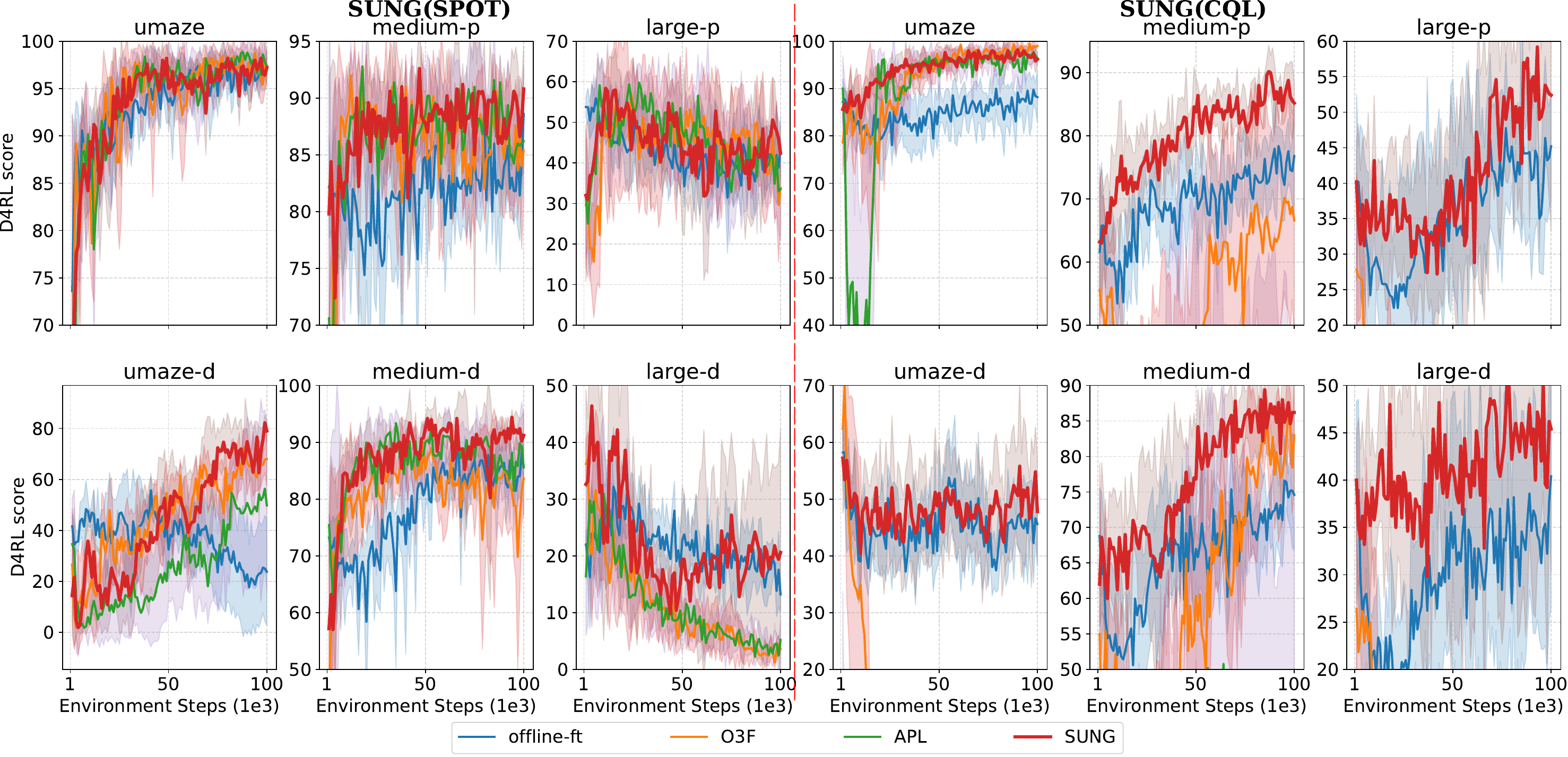}
    \caption{Learning curves of SUNG, when combined with SPOT and CQL, for 100K environment steps on \textbf{AntMaze} tasks. We also include three best-performing baselines, offline-ft, O3F and APL, for comparison. p = play, d = diverse. We report the mean D4RL score and standard deviation over 5 training seeds with 100 evaluation episodes each.}
    \label{fig:plot-antmaze}
\end{figure*}

We display the learning curves of SUNG when combined with TD3+BC and CQL for MuJoCo domains in Fig. \ref{fig:plot}. Besides, we display the learning curves of SUNG when combined with SPOT and CQL for AntMaze domains in Fig. \ref{fig:plot-antmaze}. In most settings, we can observe that SUNG surpasses the previous state-of-the-art methods in terms of both final performance and sample efficiency.

\section{Additional Empirical Analyses}

\subsection{Analyses on Other Uncertainty Quantification Method}

In this subsection, we examine an alternative uncertainty quantification method, Bayesian ensembling \cite{be-1,be-2}, which estimates epistemic uncertainty by computing the predictive variance across an ensemble of neural networks, each representing a posterior sample obtained via Markov Chain Monte Carlo (MCMC). Specifically, we train CQL \cite{cql} with 10 Q networks for offline pretraining, denoted as CQL-10. Then, we consider two variants of SUNG: \textbf{(1) SUNG w/ VAE} and \textbf{(2) SUNG w/ BE}. These variants differ in their uncertainty quantification methods. The former employs a VAE for state-action density estimation, while the latter uses the variance of 10 Q functions to estimate epistemic uncertainty. As demonstrated in Table \ref{tab:be},  the proposed VAE-based uncertainty quantification method can attain comparable performance to using the variance of 10 Q functions, which would significantly increase computational costs. Furthermore, the results also demonstrate the potential of the proposed SUNG built on top of other uncertainty quantification methods.

\begin{table}[tbp]
\caption{Comparison of the averaged D4RL score on MuJoCo tasks with \textbf{CQL-10} as the offline RL backbone method. We report the mean and standard deviation over 5 seeds.}
\begin{center}
\begin{tabular}{l|ccc}
\toprule
\hline
\multicolumn{1}{c|}{} & \textbf{CQL-10} & \textbf{\begin{tabular}[c]{@{}c@{}}SUNG\\ w/ VAE\end{tabular}} & \textbf{\begin{tabular}[c]{@{}c@{}}SUNG\\ w/ BE\end{tabular}} \\ \hline
\textbf{halfcheetah-m-v2} & 55.0 & 96.6\scalebox{0.8}{$\pm$1.0} & \textbf{98.8\scalebox{0.8}{$\pm$1.3}} \\
\textbf{hopper-m-v2} & 66.9 & 111.4\scalebox{0.8}{$\pm$0.8} & \textbf{112.5\scalebox{0.8}{$\pm$1.1}} \\
\textbf{walker2d-m-v2} & 83.2 & \textbf{113.8\scalebox{0.8}{$\pm$2.6}} & 110.2\scalebox{0.8}{$\pm$9.5} \\ \hline
\textbf{halfcheetah-m-r-v2} & 52.6 & 92.5\scalebox{0.8}{$\pm$0.5} & \textbf{96.6\scalebox{0.8}{$\pm$1.0}} \\
\textbf{hopper-m-r-v2} & 102.3 & 100.7\scalebox{0.8}{$\pm$13.6} & \textbf{107.4\scalebox{0.8}{$\pm$4.8}} \\
\textbf{walker2d-m-r-v2} & 82.1 & 105.1\scalebox{0.8}{$\pm$14.6} & \textbf{113.3\scalebox{0.8}{$\pm$16.8}} \\ \hline
\textbf{Total} & 501.0 & 620.0 & \textbf{638.8} \\ \hline
\bottomrule
\end{tabular}
\end{center}
\label{tab:be}
\end{table}

\subsection{Additional Ablation Studies} 
We consider two more ablation settings:
\begin{itemize}
    \item \textbf{Adp. Exploitation w/ Rand.:} We construct the OOD state-action pair set $\mathcal{D}_{\mathrm{OOD}}$ by randomly sampling from the mini-batch instead of using the proposed OOD sample identifier.
    \item \textbf{OORB w/o Offline Data:} We do not utilize transitions from offline dataset for online finetuning by removing offline data from OORB.
\end{itemize}

\begin{figure*}[h]
    \begin{minipage}[t]{0.48\linewidth}
    \centering
    \includegraphics[width=\linewidth]{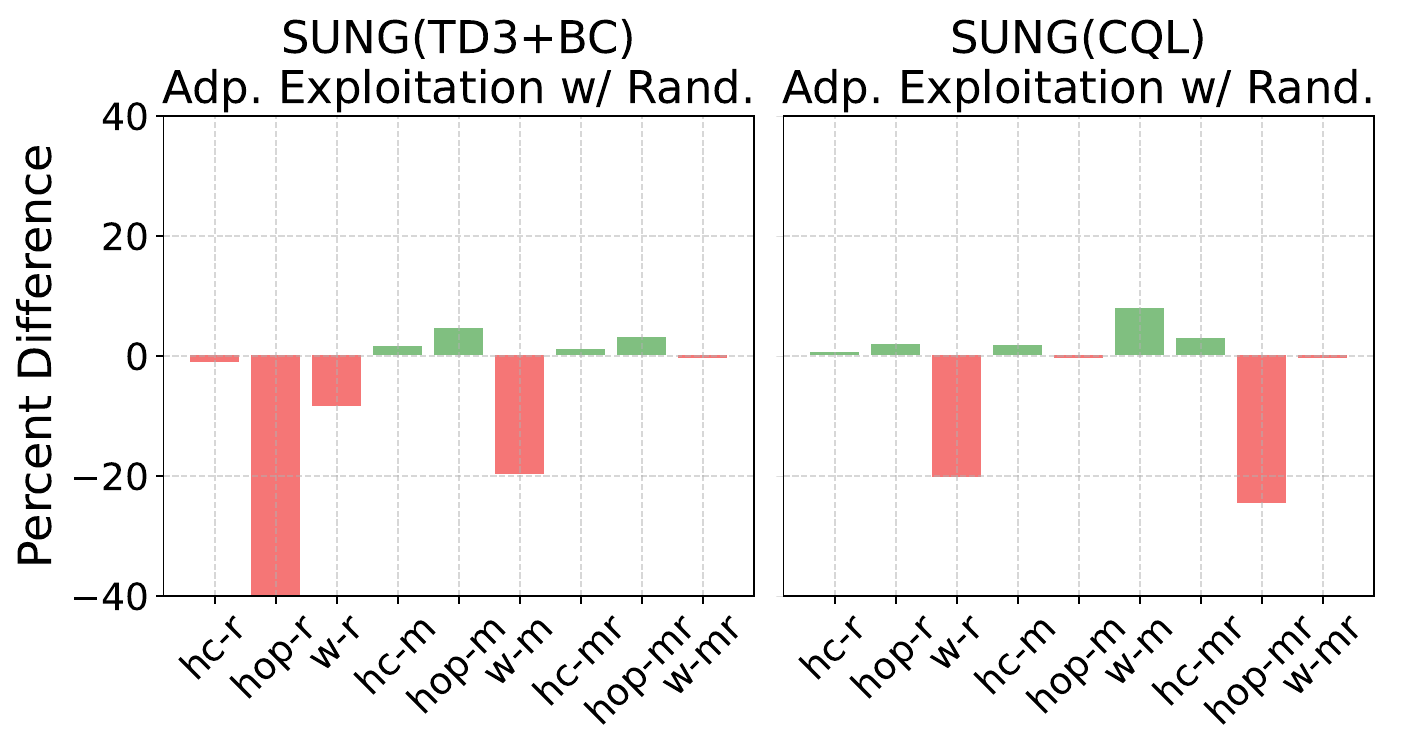}
    \caption{Performance difference of the performance of random strategy for OOD sample identification, compared with the full algorithm.}
    \label{fig:re-ablation}
    \end{minipage}
    \hfill
    \begin{minipage}[t]{0.48\linewidth}
    \centering
    \includegraphics[width=\linewidth]{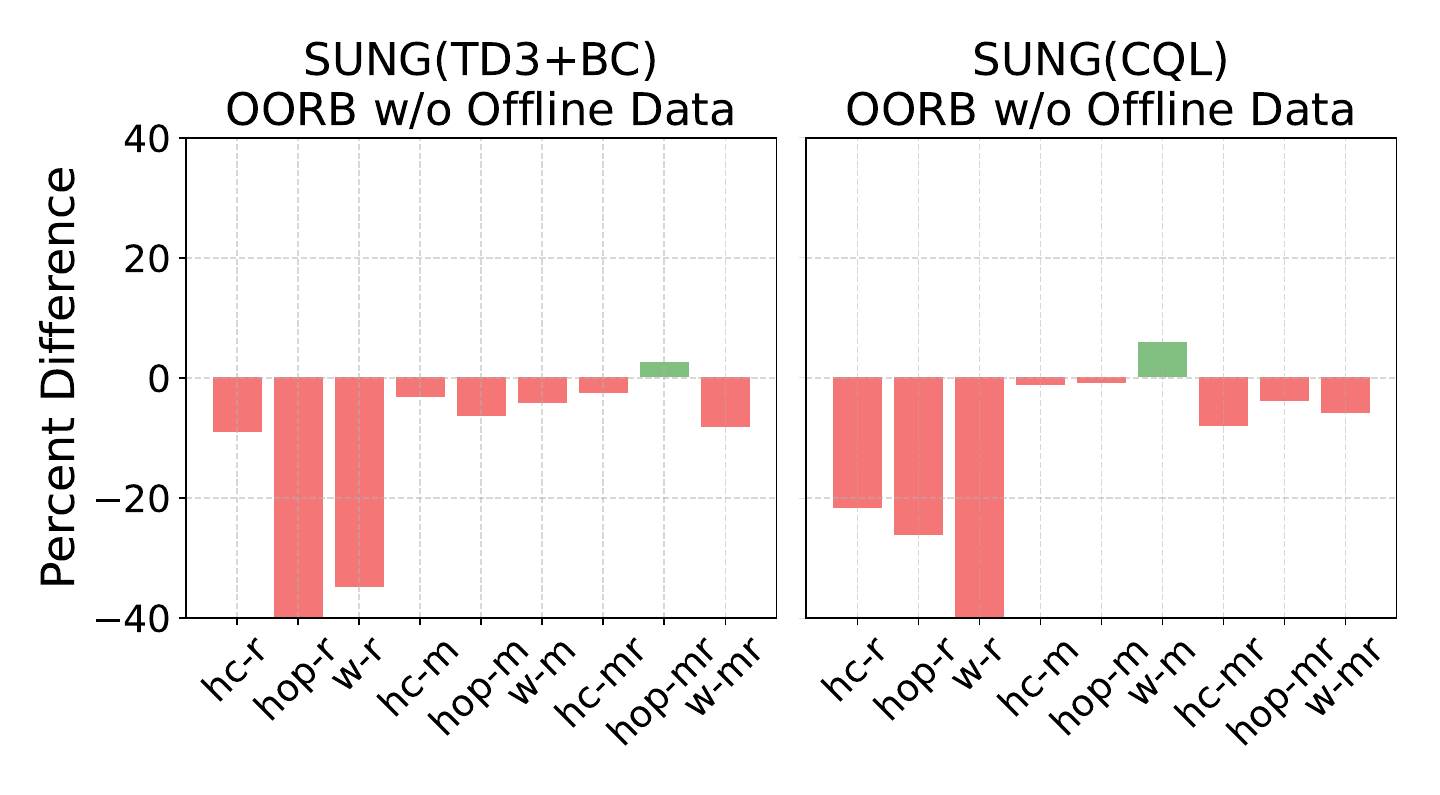}
    \caption{Performance difference of the performance without offline data, compared with the full algorithm.}
    \label{fig:od-ablation}
    \end{minipage}
\end{figure*}

We present the results of the two ablation studies above in Fig. \ref{fig:re-ablation} and Fig. \ref{fig:od-ablation}, respectively. As shown in Fig. \ref{fig:re-ablation}, we can find when we use a random sampling strategy for OOD sample identification, the performance on most settings deteriorates. As shown in Fig. \ref{fig:od-ablation}, we observe that the removal of offline data significantly hurts the offline-to-online performance. This is because online finetuning can benefit from reusing the diverse behaviours contained in the offline dataset to avoid overfitting. 

\subsection{Hyper-parameter Analysis on Softmax Temperature $\alpha$}
\begin{figure*}[h]
    \centering
    \includegraphics[width=0.8\linewidth]{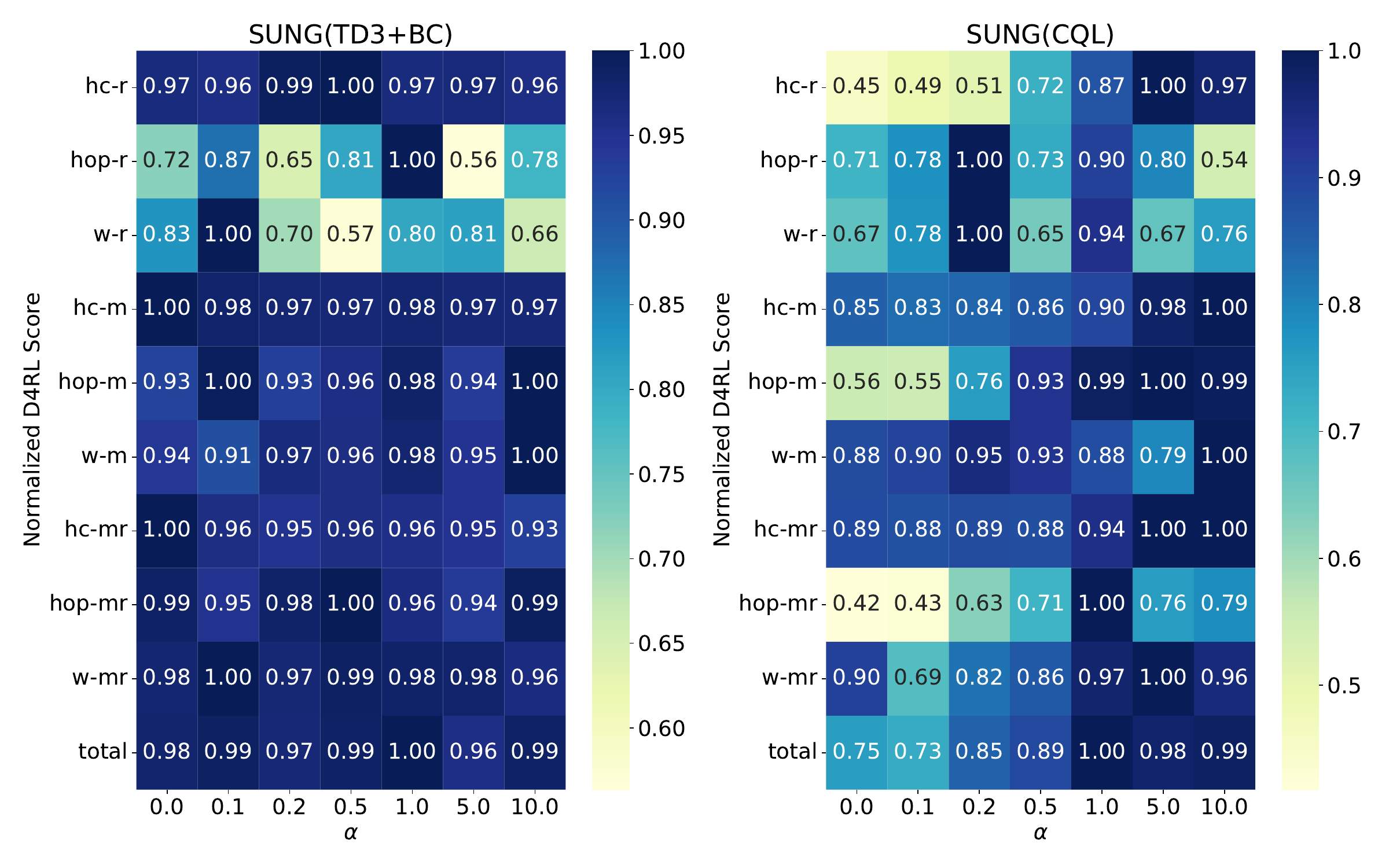}
    \caption{Heatmap for performance of SUNG with different softmax temperature $\alpha$. The results of each setting are normalized with maximum normalization for better visualization. hc = HalfCheetah, hop = Hopper, w = Walker, r = random, m = medium, mr = medium-replay.}
    \label{fig:alpha}
\end{figure*}

Here, we give a hyper-parameter analysis of the softmax temperature $\alpha$ in Eq. (10). Specifically, we investigate the choice of $\alpha \in \{0.0, 0.1, 0.2, 0.5, 1.0, 5.0, 10.0\}$ for MuJoCo domains in Fig. \ref{fig:alpha}. Note that $\alpha=0$ can be regarded as an ablation variant with the greedy selection strategy. As shown in the figure, we can find that $\alpha$ is a relatively insensitive hyper-parameter for the offline-to-online performance, and that the best-performing choice of $\alpha$ varies among different settings. Thus, though a careful tuning on $\alpha$ will bring more offline-to-online improvement, we set $\alpha=1.0$ in all the experimental result throughout the paper to guarantee the simplicity of the proposed framework. Furthermore, when we set $\alpha=0$, the performance drastically decreases, highlighting the necessity of the randomness for sampling in the exploration policy.

\subsection{Hyper-parameter Analysis on OORB Sampling Probability $p_{\mathrm{OORB}}$} 
\begin{figure*}[h]
    \centering
    \includegraphics[width=0.8\linewidth]{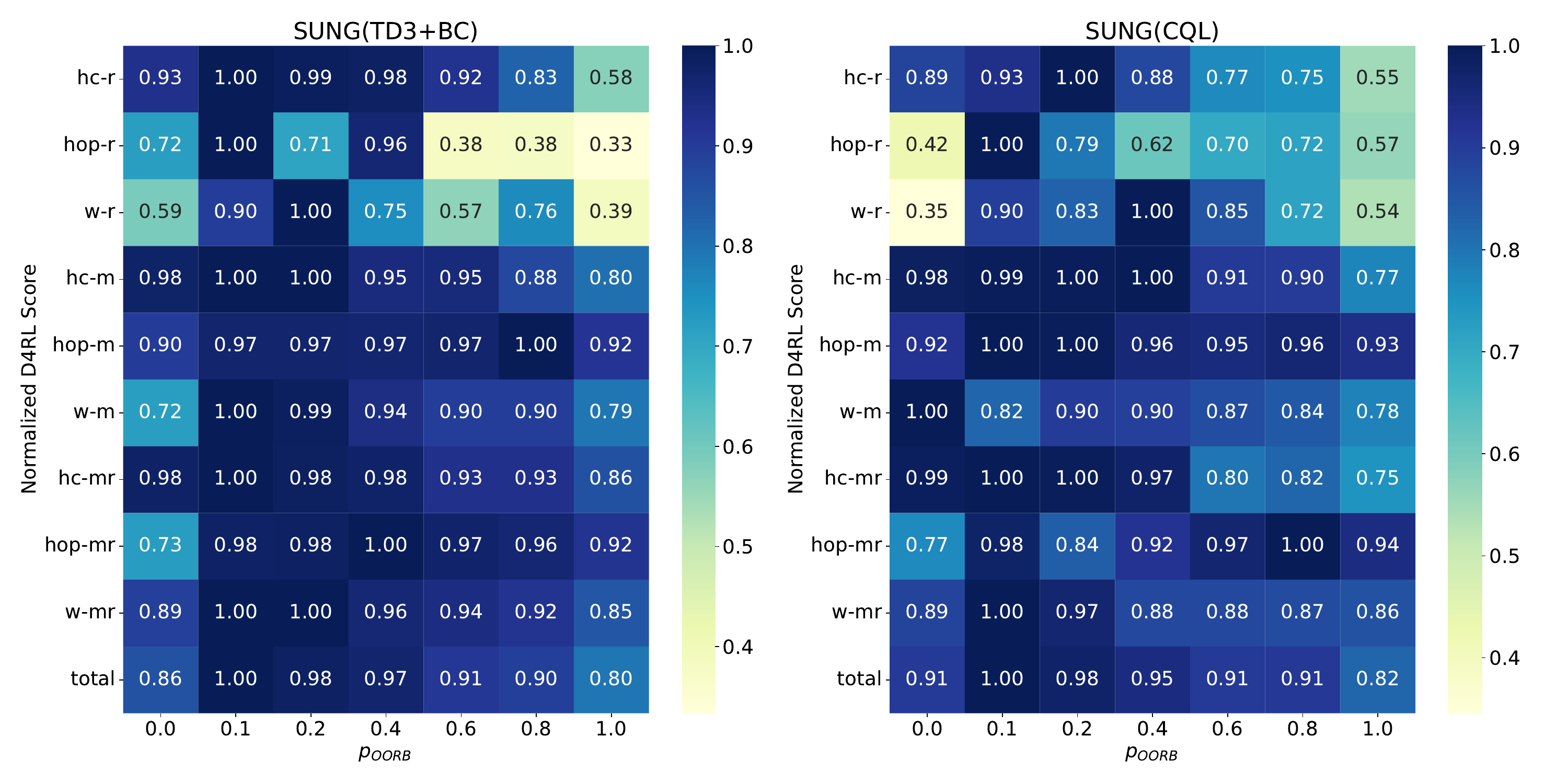}
    \caption{Heatmap for performance of SUNG with different OORB sampling probability $p_{\mathrm{OORB}}$. The results of each setting are normalized with maximum normalization for better visualization. hc = HalfCheetah, hop = Hopper, w = Walker, r = random, m = medium, mr = medium-replay.}
    \label{fig:pOORB}
\end{figure*}

Furthermore, we give a hyper-parameter analysis on $p_{\mathrm{OORB}}$ for the simple-yet-effective OORB. Specifically, we investigate the choice of OORB sampling probability $p_{\mathrm{OORB}} \in \{0.0, 0.1, 0.2, 0.4, 0.6, 0.8, 1.0\}$ for MuJoCo domains in Fig. \ref{fig:pOORB}. Note that a higher probability $p_{\mathrm{OORB}}$ indicates more reuse of the offline dataset during policy learning. From the results, we can observe that $p_{\mathrm{OORB}}=0.1$ performs the best when combined with either TD3+BC and CQL, while either setting $p_{\mathrm{OORB}}$ too conservatively (i.e., $p_{\mathrm{OORB}}=0$) or too adversely (i.e., $p_{\mathrm{OORB}}=1.0$) hurts the performance. Offline data can prevent agents from prematurely converging to sub-optimal policies due to the potential data diversity, while online data can stabilize training and accelerate convergence. Thus, it is crucial for sample-efficient offline-to-online RL algorithms to incorporate both offline and online data during policy learning.

\end{document}